\documentclass[11pt]{article}
\usepackage[margin=2.5cm]{geometry}
\usepackage{graphicx}
\usepackage{subcaption}
\usepackage{natbib}
\usepackage[hidelinks]{hyperref}
\usepackage{amsmath}
\usepackage{amsfonts}
\newcommand{\mbeq}{\overset{!}{=}}

\emergencystretch=\maxdimen
\title{Discovering dynamical laws for speech gestures}
\author{Sam Kirkham}
\date{} 

\begin{document}
\maketitle

\abstract{\noindent A fundamental challenge in the cognitive sciences is discovering the dynamics that govern behaviour. Take the example of spoken language, which is characterised by a highly variable and complex set of physical movements that map onto the small set of cognitive units that comprise language. What are the fundamental dynamical principles behind the movements that structure speech production? In this study, we discover models in the form of symbolic equations that govern articulatory gestures during speech. A sparse symbolic regression algorithm is used to discover models from kinematic data on the tongue and lips. We explore these candidate models using analytical techniques and numerical simulations, and find that a second-order linear model achieves high levels of accuracy, but a nonlinear force is required to properly model articulatory dynamics in approximately one third of cases. This supports the proposal that an autonomous, nonlinear, second-order differential equation is a viable dynamical law for articulatory gestures in speech. We conclude by identifying future opportunities and obstacles in data-driven model discovery and outline prospects for discovering the dynamical principles that govern language, brain and behaviour.}

\section{Introduction}

A longstanding goal in the cognitive sciences is the development of models that capture the dynamics of mind and motion. In the seventeenth-century, Newton proposed the fundamental laws of motion and gravitation, which synthesized a diverse range of observations into a unified mathematical framework. This identification of common mathematical laws has come to represent a core goal of modern science \citep[e.g.][]{newton1687, noether1918, anderson1972}, but it has long been acknowledged that the dynamics of living systems are structured by other laws beyond fundamental physics \citep{schrodinger1944}. For example, within decades of Newton's discoveries, a new paradigm emerged that was focused on discovering fundamental laws behind human behaviour and cognition \citep{hume1739}, thus laying the foundations for contemporary cognitive science. Two central questions in this line of inquiry are: (1) what are the principles that govern behaviour and cognition? (2) can these principles be expressed as mathematical laws?

 An exemplary distillation of this challenge is the case of spoken language. Spoken (and signed) languages involve mapping a set of high-dimensional continuous physical movements to a low-dimensional set of discrete tasks that comprise the combinatorial units of language. A fundamental issue concerns the relation between these two components: the qualitative aspects of phonological knowledge and their physical realization. One view proposes a translation between discrete symbolic units and continuous physical properties, such that phonetic realization is a matter of translation or an `interface' between symbolic and physical domains \citep{chomsky-halle1968, keating1990, guenther2016, turk-shattuck-hufnagel2020}. An alternative view holds that the relation between discrete and continuous aspects of phonological cognition can be explained using the language of nonlinear dynamics \citep{browman-goldstein1986}. In this sense, phonetics and phonology are isomorphic, rather than separate modules requiring translation, and can be cast as intrinsically linked elements within a single dynamical system \citep{kelso-etal1986, browman-goldstein1992, gafos2006}.
  
 The dynamical view of phonology emerges from a broader perspective in \textit{dynamical systems theory}, which views the world through the mathematical language of change: differential equations. Dynamical thinking has a long history in the cognitive sciences \citep[e.g.][]{fowler1980, smolensky1988, kelso1995, port-vangelder1998} and is typically set in opposition to highly modular models of mind (e.g. \citealt{turing1950, fodor1975}). The overarching goal is to identify the appropriate dynamical laws that govern brain and behaviour across task-specific domains, including the relations between microscopic and macroscopic scales. For example, a key analytic concept is identifying the appropriate `order parameters' \citep{haken1977} or low-dimensional variables that govern qualitative states in the dynamical system \citep{haken-etal1985}. This does not necessarily mean doing away with symbolic representations; e.g. \citet{gafos2006} outlines a cogent theory of the relation between dynamics and phonological grammar that re-casts the `interface' as a `dynamic linkage'. Indeed, many dynamical theories maintain representations and symbolic systems, but the primary hypothesis is that cognitive agents are dynamical systems that can be cast in terms of states, paths and flows, rather than modules, computations and translations \citep{vangelder1998}.\footnote{ For excellent introductions to dynamical systems and nonlinear dynamics, see \citet{strogatz2015} (a mathematical introduction); \citet{abraham-shaw1992} (a visual introduction); \citet{vangelder1998} (a perspective from the cognitive sciences); \citet{kelso1995} (coordination), \citep{haken1977} (synergetics); \citet{gafos2006} (phonology); and \citet{tilsen2019b} (an impressive application of dynamical theory to syntactic structures). For arguments against computational metaphors of mind see \citet{gibson1979, carello-etal1984, spivey2007, chemero2009, barrett2011}.}

Given the immense complexity of human cognition, accompanied by our highly variable and noisy behavioural measurements, how do we begin to identify appropriate dynamical models of brain and behaviour? In this study, we focus on developing models of articulatory trajectories -- or `gestures' -- in the human vocal tract. This is a well-studied and tractable problem, as it is possible to collect lots of high-quality data on the movements of speech articulators, which we can use to  derive the lawful regularities that govern such movements. We approach this problem as one of data-driven model discovery and, in doing so, leverage recent developments in physics-informed machine learning and symbolic regression, which allow us to learn simple and interpretable dynamical equations directly from large datasets. The outcome is a small set of candidate models, from which new predictions can be generated and tested. Before outlining this approach in more detail, we first review extant dynamical models of speech, before motivating the search for new models.

\subsection{Dynamical models of speech}
\label{subsec:dynamical_models}

In their article `The dynamical perspective on speech production', \citet{kelso-etal1986} argue that the task for a dynamical model of speech is `less one of translating a ``timeless" symbolic representation into space-time articulatory behavior, as it is one of relating dynamics that operate on different intrinsic time scales'. The dynamical timescales involved in speech production span macroscopic forms of organization (syllables, words, prosodic structure) to microscopic physical activity (articulators, biomechanics, neural dynamics), and may even expand to contextualize speech production in terms of a larger agent-environment dynamical system (perception-action dynamics; interactions with other speakers; etc). The most thoroughly worked out theory of dynamical systems in the study of spoken language is Articulatory Phonology / Task Dynamics \citep{fowler1980, saltzman-munhall1989, browman-goldstein1992, gafos-benus2006, tilsen2016, iskarous2017}, henceforth AP/TD. In this theory, the fundamental unit is the \textit{gesture}, an abstract force acting on the vocal tract that drives it from its current state towards a new target state \citep{browman-goldstein1992}. To this end, the gesture is both a model of articulatory motion and a model of the `cognitive control of abstract linguistic units' \citep[154]{byrd-saltzman2003}.

\begin{figure}[h]
\centering
\includegraphics[scale=0.05]{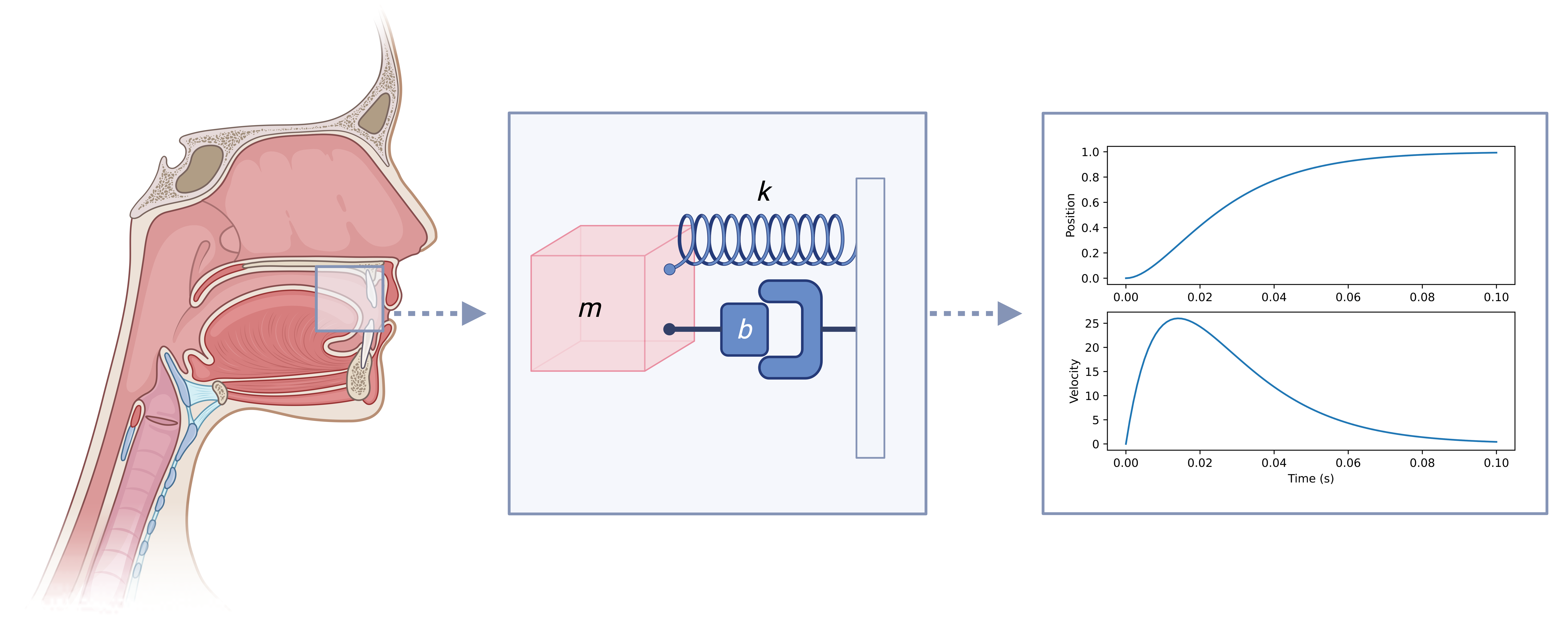}
\caption{The damped-mass spring model is a model of vocal tract dynamics. The left diagram shows a midsagittal view of the vocal tract, with a box centred on the Tongue Tip task space. The middle diagram shows a physical damped mass-spring system representing the forces that act on the Tongue Tip gesture, where $m$ is a mass, $k$ is spring stiffness, and $b$ is the strength of the damping force. The right diagram shows simulated trajectories from the damped mass-spring model, with the Tongue Tip moving from a low to a high position (arbitrary units).}
\label{fig:vt_spring_model}
\end{figure}

A common mathematical model of speech gestures in AP/TD is the damped mass-spring model, which is visualized in Figure \ref{fig:vt_spring_model} and captures the dynamics of forces on the vocal tract \citep{saltzman-munhall1989}. The model approximates an idealized physical system, with a physical mass attached to a spring, plus a damper or shock absorber. Variants of this model have been well-studied since the seventeenth century \citep{hooke1678} and the model can be easily understood by way of its physical properties, such as the analytical relationship between spring damping/stiffness and the system's position, velocity and acceleration. These insights can then be generalized to the dynamics of vocal tract tasks during speech, allowing us to advance a specific and testable model of speech dynamics based on a well-understood physical analogy.

The dynamics of the speech gesture can be explicitly formalised as a second-order, critically damped harmonic oscillator, as in (\ref{eq:sm89}):

\begin{equation}
m\ddot{x} + b\dot{x} + kx = 0
\label{eq:sm89}
\end{equation}

In the above equation, $x$ represents the state of the system, such as the current state of an articulatory variable, while $\dot{x}$ and $\ddot{x}$ respectively represent the velocity and acceleration of $x$. The system's mass $m$ is typically set to $m = 1$ and $k$ is a stiffness constant. The damping constant is $b = 2\sqrt{mk}$ in critically damped versions of the model. Equation (\ref{eq:sm89}) specifies the system's equilibrium or target position as zero, but non-zero targets can be introduced by adding a target parameter (which we denote $T$) to the $kx$ term as in (\ref{eq:sm89_T}). We assume that $T$ is implicit in any subsequent formulations of these equations.

\begin{equation}
m\ddot{x} + b\dot{x} + k(x-T) = 0
\label{eq:sm89_T}
\end{equation}

This equation defines the relationships between parameters as invariant over an instance of the system, where the system evolves until the value of $T$ is reached. Research on the neural encoding of speech movements has identified signatures of critically damped oscillations in the neural populations of ventral sensorimotor cortex, supporting a neural basis for the kinds of oscillatory models used in AP/TD \citep{chartier-etal2018}.

The state of a gestural system evolves towards its equilibrium or target position, after which the target parameter changes and the system evolves towards a new state. This affords movement between different vocal tract postures, which is a fundamental characteristic of speech. Change in parameters requires a notion of gestural activation; if we cast gestural activation as a function of time $a(t)$ then we can transform (\ref{eq:sm89}) into (\ref{eq:sm89_act}).

\begin{equation}
m\ddot{x} + a(t)[b\dot{x} + kx] = 0
\label{eq:sm89_act}
\end{equation}

In the standard model of \citet{saltzman-munhall1989}, $a(t)$ in Equation (\ref{eq:sm89_act}) corresponds to the rectangular pulse function in (\ref{eq:rect}). This means that gestural parameters change instantaneously at the point of gestural activation $a(t) = 1$ and remain constant until activation ends $a(t) = 0$, where activation is bounded by the temporal interval $[t_{a}, t_{b}]$.

\begin{equation}
a(t) = 
\begin{cases} 
    1, & t \in [t_{a}, t_{b}],\\ 
    0, & \text{otherwise}
\end{cases}
\label{eq:rect}
\end{equation}

Equation (\ref{eq:sm89_act}) is a critically damped harmonic oscillator with step activation. It is well-established that this model fails to capture many empirical characteristics of speech movements. This includes overly short time-to-peak velocity and highly asymmetric velocity trajectories compared with those seen in empirical data \citep{byrd-saltzman1998, sorensen-gafos2016}. An alternative approach is to introduce time-varying activation \citep{byrd-saltzman1998}. Equation (\ref{eq:sine}) is an example of a continuous activation function from \citet{kroger-etal1995}.

\begin{equation}
a(t) = 
\begin{cases} 
    0, & t < t_{a}\\ 
    \sin\left(\frac{2\pi(t-t_{a})}{4(t_{b}-t_{a})}\right), & t_{a} \leq t < t_{b}\\
    1, & t_{b} \leq t < t_{c}\\
    \sin\left(\frac{2\pi(t-t_{d})}{4(t_{c}-t_{d})}\right), & t_{c} \leq t < t_{d}\\
    0, & t > t_{d}
\end{cases}
\label{eq:sine}
\end{equation}

The activation function in (\ref{eq:sine}) defines a quarter sine rise over $[t_{a}, t_{b})$, steady activation over $[t_{b}, t_{c})$, a quarter sine fall over  $[t_{c}, t_{d})$, and zero activation outside the range $[t_{a}, t_{d}]$. The ramped activation function corrects for the short time-to-peak velocities and velocity asymmetries of step activation \citep{byrd-saltzman1998}. In both step and continuous activation, gestures are active when they exceed a threshold of zero (with possible values of 0 or 1). An alternative view casts gestures as continuously active \citep{tilsen2020}, which we discuss in Section \ref{sec:discussion}.

Common to models of continuous gestural activation is that the system is explicitly time-dependent (i.e. non-autonomous) during activation. An alternative approach is to re-formulate the dynamical equations that govern gestural dynamics, instead of using a more complex activation function. For example, \citet{sorensen-gafos2016, sorensen-gafos2023} introduce a cubic term in (\ref{sg16}), but retain the  rectangular pulse function in (\ref{eq:rect}) for the activation variable $a(t)$. Note that the $m$ term has been omitted from (\ref{sg16}) and all subsequent related equations because it is conventional to define $m = 1$ (although see \citet{simko-cummins2010} for a task dynamic model where different gestures are defined over different masses).

\begin{equation}
\label{sg16}
\ddot{x} + a(t)[b\dot{x} + kx - dx^3] = 0
\end{equation}

\begin{figure}[h]
\centering
\includegraphics[scale=0.6]{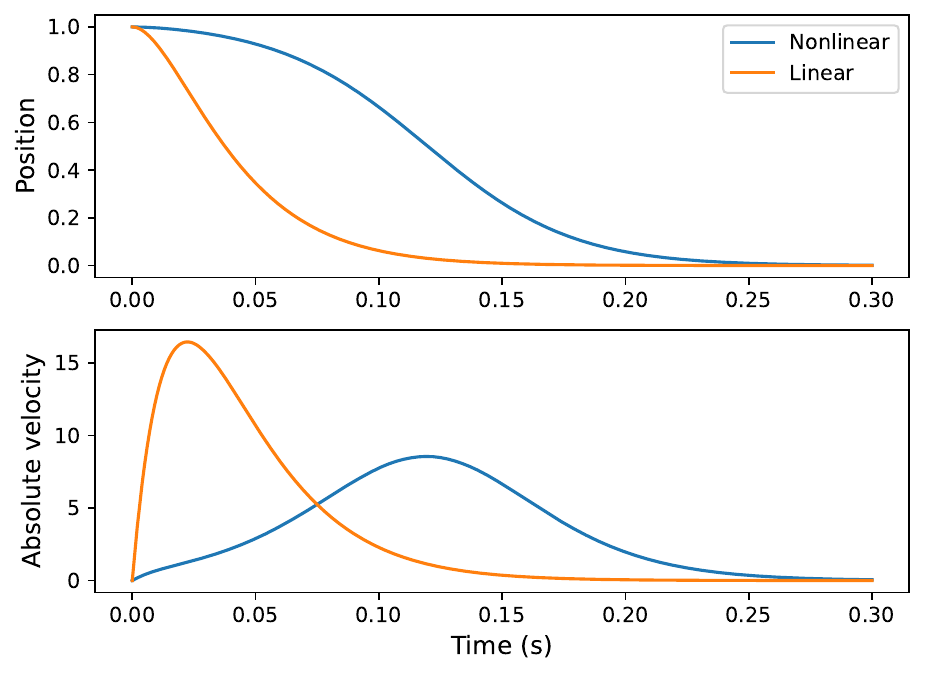}
\caption{Position and absolute velocity trajectories simulated using a linear damped mass-spring model and a nonlinear (cubic) damped mass-spring model. In both cases, $x_{0} = 1, \dot{x}_{0} = 0, T = 0, k = 2000, b = 2\sqrt k$. The nonlinear cubic coefficient is $d = 0.95k$.}
\label{fig:sm89_sg16}
\end{figure}

The cubic term in (\ref{sg16}) has the effect of acting as a nonlinear restoring force on the spring in the mass-spring model, with $d$ governing the strength of the nonlinear force. This means that the effect of the nonlinear force is greater with larger movement displacement, thereby reproducing observed nonlinear empirical relations between movement amplitude and peak velocity \citep{ostry-munhall1985}. The model also generates the symmetrical velocity profiles observed in empirical data, while retaining the step-function activation, which yields constant forcing during periods of gestural activation. The effects of this can be seen in Figure \ref{fig:sm89_sg16}, which compares the linear and nonlinear task dynamic models under step activation.

The prevailing philosophy in the above research seeks minimal and autonomous models that capture fundamental dynamics without recourse to extrinsic timing mechanisms \citep{sorensen-gafos2016, iskarous-etal2024}. But we should be open to the possibility that further good models may be possible. For example, \citet{turk-shattuck-hufnagel2020} propose a General Tau model for articulatory movements, as part of a broader symbolic theory of phonology. Their model fits position and velocity trajectories better than the nonlinear dynamical model in \citet{sorensen-gafos2016}, although a comprehensive evaluation of how different models fit acceleration data is yet to be established \citep{sorensen-gafos2016}. There are also reasons beyond quantitative fit to search for new models, such as the trade-off between simplicity and accuracy, and the value of interpretable parameters. Standard models may also not capture all kinds of speech and may require modifications or extensions to reproduce phenomena observed in disordered speech \citep{mucke-etal2024}, speech development \citep{abakarova-etal2022}, different languages \citep{geissler-nellakra2024}, and so on. Finally, discovering new models is a core part of evaluating the success of existing models; if we can develop better models then this would represent a major advance, but if extant models prove more successful than the discovered models then this is also an important empirical finding.

\section{Discovering dynamical models from data}
\label{sec:discovering_models}

\subsection{What does a good model look like?}

\begin{figure}[h]
\centering
\includegraphics[scale=0.7]{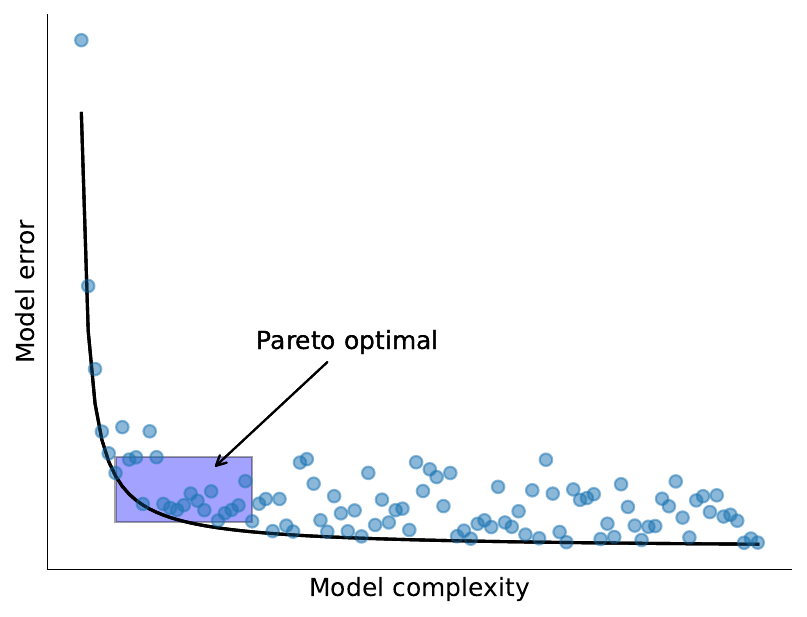}
\caption{A Pareto curve showing the schematized relationship between model accuracy and model complexity, after \citet{brunton-kutz2022}. An ideal model occupies the Pareto optimal space, which strikes a balance between accuracy and simplicity.}
\label{fig:pareto}
\end{figure}

There is a near infinite number of \textit{potential} models that could have generated a spoken utterance, but the number of \textit{good} models is likely few in number. How do we know which models are good? A typical approach in model discovery is to strike a balance between two characteristics -- parsimony and accuracy -- which is schematized as complexity versus error in Figure \ref{fig:pareto}.

A maximally parsimonious model is a very simple model with as few terms as possible in the equation (top left of Figure \ref{fig:pareto}). A parsimonious model is highly interpretable, because we have a small number of model terms that make clear predictions. The downside is that those predictions are likely to be inaccurate, as they are a poor fit with the real world due to insufficient complexity. At the other extreme, a maximally accurate model would have a very high number of terms, which offer immense flexibility in fitting the model to highly variable data (bottom right of Figure \ref{fig:pareto}). An example of this would be a deep neural network, which can have upwards of many hundreds of parameters. The downside is that such a model is likely to be incredibly complex and uninterpretable, meaning we learn little about the system's fundamental dynamics. Our aim is to discover models that fit into the `Pareto optimal' space in Figure \ref{fig:pareto}, representing models that are simple but show high accuracy. The following section addresses the conceptual and technical solution to this approach.

\subsection{Sparse identification of nonlinear dynamics}
\label{subsec:sindy}

A popular technique for data-driven model discovery in physics and engineering is the class of SINDy (Sparse Identification of Nonlinear Dynamics) methods that have emerged over the past decade \citep{brunton-etal2016}. SINDy is based on the principles of symbolic regression \citep{schmidt-lipson2009}, which aims to approximate an unknown function from some data $X, \dot{X}$ as a combination of nonlinear functions.

\begin{equation}
\dot{X} = \Theta(X)
\end{equation}

where $\Theta(X)$ is a feature library composed of an arbitrary number of mathematical functions.

\begin{equation}
\Theta(X) = [1 X X^2 X^3 \dotsc \sin X \cos X]
\end{equation}

The aim is to discover coefficients for the functions in $\Theta(X)$ that explain variation in $\dot{X}$. A known problem of symbolic regression is that the above procedure will produce many non-zero coefficients, leading to a complex model that contains many terms from the feature library. Such models are likely overfitted to the data and may be more complex than desirable. How can we only retain the terms that contribute substantially to the system under study? SINDy solves this by discovering the optimal \textit{sparse} coefficient matrix $\Xi$ corresponding to the functions in $\Theta(X)$.

\begin{equation}
\Xi = [\xi_{1} \xi_{2}  \xi_{3}  \dotsc \xi{n}]
\end{equation}

The aim is to discover a coefficient matrix $\Xi$ that provides an excellent fit to $\dot{X}$ while being as sparse as possible; i.e. containing the smallest number of terms required to produce a good fit. This can be cast as Equation (\ref{eq:sindy}).

\begin{equation}
\dot{X} = \Theta(X)\Xi
\label{eq:sindy}
\end{equation}

A range of sparsity-promoting algorithms exist to solve this problem, two of which are reviewed in Section \ref{subsec:comp_impl}. The sparsity-promoting characteristics of SINDy mean that a symbolic model derived from data often performs better at estimating parameters from new data than a higher-dimensional neural network. Subsequent research has expanded the range of optimization methods for diverse problems in physics, biology and engineering, providing a valuable toolkit for data-driven model discovery \citep[e.g.][]{kaiser-etal2018, champion-etal2020}.

SINDy holds great promise for learning interpretable and parsimonious models of brain and behaviour. \citet{dale-bhat2018} review extant approaches to equation discovery in the cognitive sciences and outline some prospects and challenges for applying SINDy to social and cognitive systems. They highlight difficulties associated with discovering models from noisy and continuously changing systems and pose a number of valuable recommendations for future research. To date, however, there have only been very limited applications of SINDy in cognitive science. In terms of related methods, \citet{iskarous2016} shows how least squares regression can be used to estimate dynamical models of action and perception from data, while \citet{nalepka-etal2019} model human multi-agent activity in games using dynamical motor primitives. Specific applications of SINDy include discovering mechanisms of human learning from experimental data \citep{lafollette-etal2024} and a proof-of-concept study on discovering dynamical models of speech \citep{kirkham2024}.

\subsection{An example: discovering known models from simulated data}
\label{subsec:sim_example}

We now turn to a brief illustration of model discovery using SINDy on simulated data. This outlines the conceptual steps involved in model discovery and validates that the method can discover a known model. As an example, we generate simulated data using the damped harmonic oscillator in (\ref{sm89_target}), as this is a widely used model of articulatory gestures in speech production. The equation is written with acceleration $\ddot{x}$ on the left-hand side and damping $b\dot{x}$ and stiffness $k(x-T)$ terms on the right hand side, where $T$ is the target or equilibrium position of the system. The damping coefficient $b$ is typically defined as $b = 2\sqrt k$, which makes (\ref{sm89_target}) a critically damped harmonic oscillator that will asymptotically approach the target. Note that $\ddot{x}$ is typically written as $m\ddot{x}$, but as $m = 1$ it is omitted from here onwards.

\begin{equation}
\label{sm89_target}
\ddot{x} = -b\dot{x} -k(x-T)
\end{equation}

We solve a trajectory from this equation based on a set of initial position $x_{0}$ and velocity $\dot{x}_{0}$ values, and a stiffness $k$ and target $T$ value, all of which remain invariant over the time-course of the simulation. Numerical solutions were computed using Python's \texttt{scipy.integrate.solve\_ivp} function \citep{SciPy-NMeth2020}, with a Runge-Kutta method of order 5(4) and a timestep of $\Delta t = 0.001$. Position and velocity trajectories were simulated using initial conditions $x_{0} = 1$, $\dot{x} = 0$, and parameter values $k = 2000$, $b = 2\sqrt k = 89.44$ and $T = 0.2$.

\begin{figure}[h]
\centering
\includegraphics[scale=0.6]{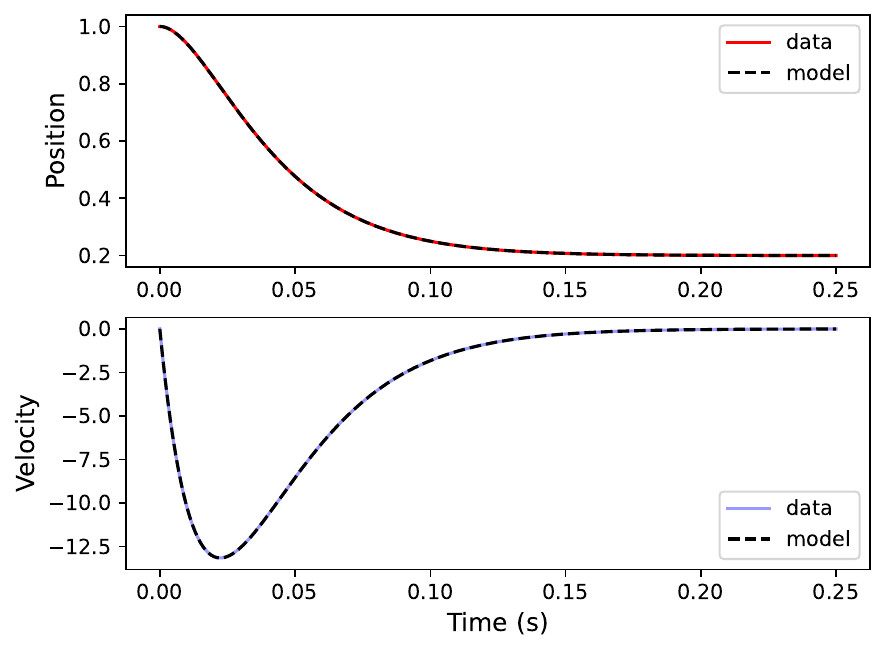}
\caption{Simulated position and velocity trajectories plotted against SINDy model predictions.}
\label{fig:sim-data}
\end{figure}

We pass the simulated position and velocity trajectories to a SINDy algorithm, with a first-degree polynomial library and a coefficient threshold of 0.1. Figure \ref{fig:sim-data} shows a predicted trajectory from the discovered model plotted on top of the data, with no visible differences between data and prediction, resulting in a fit of $R^2 = 1.00$. The discovered symbolic equation is $\ddot{x} = -b\dot{x} - kx + kT$ and we can rearrange terms to get $\ddot{x} = -b\dot{x} -k(x-T)$, which is the original equation that generated the simulated data. The discovered coefficients are $kT = 400.299$, $k = 2001.31$, $b = 89.47$. If we calculate $T = \frac{kT}{k}$ then $T = 0.2$, showing that all discovered coefficients are within 0.1\% of their true values. This shows that SINDy is able to accurately discover symbolic models and coefficient values from data; \citet{kirkham2024} shows that the same model can be accurately discovered even when considerable noise is added to simulated trajectories. The rest of this article applies this method to empirical data on movements of the tongue and lips during speech, with the aim of discovering accurate and interpretable models of gestural dynamics.

\section{Methods}

\subsection{Data}

We report an experiment demonstrating the use of the SINDy framework for discovering the articulatory dynamics of continuous speech. We use data from the X-Ray Microbeam (XRMB) corpus \citep{westbury1994}, which contains articulatory speech data from a relatively large number of speakers. The XRMB corpus contains data from 57 speakers, most of whom speak an Upper Midwest dialect of American English (32 female, 25 male), with a median age of 21. We use a subset of 48 speakers, corresponding to those who have existing forced-aligned phone-level annotations available in the database at \url{https://github.com/rsprouse/xray_microbeam_database}. The data comprise small pellets of 2.5--3.0 mm in size attached to the upper (UL) and lower (LL) lips, surface of the tongue (T1, T2, T3, T4), mandible and head. The pellets were tracked using a narrow X-ray beam at 160 Hz for T1, 80 Hz for T2, T3, T4 and LL, and 40 Hz for UL, after which all pellets were resampled at 160 Hz and translated/rotated to an anatomically-defined coordinate system. See \citet{westbury1994} for comprehensive technical documentation of the data and speaker sample.

We discover models using data from XRMB tasks \#11 and \#101, both of which feature continuous speech of varying durations. Task \#11 involved each speaker reading the following passage:

\begin{quote}
You wish to know all about my grandfather. Well, he is nearly 93 years old, yet he still thinks as swiftly as ever. He dresses himself in an old black frock coat, usually several buttons missing. A long beard clings to his chin, giving those who observe him a pronounced feeling of the utmost respect. When he speaks, his voice quivers a bit. Twice each day he plays skillfully upon a small organ.
\end{quote}

Task \#101 involved each speaker producing the following three sentences in a single recording, with a short pause between each phrase:

\begin{itemize}
  \item Elderly people are often excluded.
  \item When all else fails, use force.
  \item The dormitory is between the house and the school.
\end{itemize}

\subsection{Data processing}
\label{subsec:data_processing}

Four articulatory variables were extracted from every recording: lip aperture (LA), tongue tip (TT, based on T1), tongue dorsum (TD, based on T3), tongue root (TR, based on T4). We model gestural dynamics in one dimension, so we reduce all articulatory motions as follows. LA is the Euclidean distance between the vertical coordinates of the upper and lower lips, while TT, TD and TR are the first principal component projected from the two-dimensional $x/y$ coordinates, which was calculated separately for each speaker and each sensor \citep{birkholz-etal2011, elie-etal2023}. We focus on discovering autonomous dynamical models that assume step activation of the type in Equation (\ref{eq:rect}), which has direct consequences for our approach to gestural segmentation. All signals were first divided into inter-pause intervals, based on the forced aligner's segmentation, and then gestural segmentation was calculated on the basis of zero-crossings in the velocity signal. A gesture is defined as a velocity peak or trough bounded by two velocity zero-crossings. Some velocity trajectories have a peak that never crosses zero, meaning that such trajectories often have two or more peaks. Trajectories with two or more peaks were identified algorithmically and excluded from the analysis, which corresponds to 13.3\% (N = 2924) of trajectories. These trajectories were excluded because existing task dynamic models do not predict multiple velocity peaks per gesture. We also excluded trajectories longer than 200 ms in duration, because many of these represent passive speech movements during periods in which that articulatory variable was not actively involved in the production of a constriction. This corresponds to 24.4\% (N = 5383) of trajectories, which also includes a small number of additional trajectories that were clearly erroneous based on visual inspection.

In total, we analyse 62.3\% (N = 13,742) of the segmented trajectories from the dataset. While this is a conservative approach to data filtering, it is likely that a large number of the excluded trajectories represent uncontrolled movements, anomalies at signal edges, segmentation errors, and inaccurate velocity calculations. We nonetheless retain a very large number of target gestures that is far beyond the amount possible to check manually. In total we analyse 13,742 individual gestures across four articulatory variables (LA 3894, TT 3715, TD 3027, TR 3106). We discover models from each articulatory variable separately in order to test whether model discovery is robust across articulatory tasks.

\subsection{Computational implementation}
\label{subsec:comp_impl}

Our computational modelling is based on the Python package \texttt{pySINDy}, which is a computational framework for discovery of governing dynamical equations from data \citep{desilva-etal2020}. Each step of the model discovery procedure is explained below. Data and code for reproducing every analysis in this article is available at: \url{http://doi.org/10.5281/zenodo.15101639}.

\subsubsection{Model inputs}

This study involves the discovery of two kinds of models: (1) first-order models, where the models only depend on the position and velocity of the system; and (2) second-order models, where the system includes position, velocity and acceleration. The task dynamic models in \citet{saltzman-munhall1989} and \citet{sorensen-gafos2016} are both examples of second-order models, and it is well-known that modelling skilled movements typically requires a second-order model \citep{saltzman-kelso1987}. To provide a point of comparison, however, we also test the hypothesis that articulatory movements can be approximated by first-order models, with no information about acceleration. First-order models can accurately model other physical dimensions of speech, such as fundamental frequency contours \citep{iskarous-etal2024}, although note that these models in \citet{iskarous-etal2024} are a series of coupled agonist-antagonist equations, which are more complex than the single-order models examined here.

The target system to be modelled is the intrinsic dynamics of the speech gesture, as defined in Section \ref{subsec:data_processing}. This corresponds to an interval including the initiation of a gesture, its movement towards its target, and target achievement. The movement away from the target represents a new gesture, which is either a release gesture (e.g. returning to a rest position), or a movement towards a different target.  The input data is a position signal for first-order models, and a position signal with its associated velocity signal for second-order models. In no cases do we find substantially better model fits for opening versus closing gestures, which are instead reflected in simple parameter differences, so all models collapse across this distinction. The technical details of data processing were reported in Section \ref{subsec:data_processing}.

\subsubsection{Sequential Thresholded Least Squares (STLSQ)}

We now review algorithms that promote sparsity in the discovered models. Recall from Section \ref{subsec:sindy} that Equation (\ref{eq:sindy2}) defines the goal of model discovery, where the aim is to discover the sparse coefficient matrix $\Xi$ corresponding to features in the library $\Theta(X)$ that optimally model the time-derivative $\dot{X}$.

\begin{equation}
\dot{X} = \Theta(X)\Xi
\label{eq:sindy2}
\end{equation}

One of the simplest methods is Sequential Thresholded Least-Squares (STLSQ) \citep{brunton-etal2016}. STLSQ aims to (1) obtain a least squares solution for $\Xi$; (2) eliminate any coefficients below a pre-defined threshold; (3) iterate until the procedure converges on an optimally sparse model. This procedure can be cast as the objective function in (\ref{eq:stlsq}), where $\Xi$ is the sparse coefficient matrix to be optimized and $\alpha$ defines the $\ell_{2}$ regularization weight, which makes the problem a form of sequentially thresholded ridge regression. Note that the threshold parameter $\lambda$ is not explicitly specified in the objective function as it is applied post-hoc over repeated iterations of (\ref{eq:stlsq}).

\begin{equation}
\min\limits_{\Xi} ||\dot{X} - \Theta(X)\Xi||^2_{2} + \alpha||\Xi||^2_{2}
\label{eq:stlsq}
\end{equation}

In the present study, we use STLSQ for first-order models, because it provides an effective and simple technique for sparse model discovery. We set $\alpha = 0.05$, use a maximum of 20 optimization iterations to allow for convergence of the thresholding algorithm, and a coefficient threshold optimized to maximize $R^2$ model fit from the set $\lambda \in \{0.001, 0.01, 0.1\}$.

\subsubsection{Sparse Relaxed Regularized Regression (SR3)}

While STLSQ works well for first-order models that only involve a single time derivative (i.e. velocity), a second-order model introduces additional complexity that can be better constrained using alternative techniques. For example, when we numerically solve a second-order differential equation as in (\ref{eq:secondorder1}), we split it into two coupled first-order equations as in (\ref{eq:secondorder2}) and (\ref{eq:secondorder3}). This involves the introduction of a new variable $y$, requiring us to solve for $y$ and $\dot{y}$.

\begin{equation}
\ddot{x} = -b\dot{x} - kx
\label{eq:secondorder1}
\end{equation}

\begin{equation}
y = \dot{x}
\label{eq:secondorder2}
\end{equation}

\begin{equation}
\dot{y} = -by - kx
\label{eq:secondorder3}
\end{equation}

A SINDy algorithm will simultaneously discover two equations for a second-order model. If the equations are of the form in (\ref{eq:secondorder2}) and (\ref{eq:secondorder3}) then we can simply substitute $y$ into (\ref{eq:secondorder3}) and rearrange to obtain our second-order model for $\ddot{x}$. However, one consequence of the model discovery procedure is that SINDy will try to fit all terms in the library to both equations. In principle, this means that equation (\ref{eq:secondorder2}) could be something other than $y = \dot{x}$, such as (\ref{eq:secondorder4}). 

\begin{equation}
y = 0.97\dot{x} + 0.21x - 0.13
\label{eq:secondorder4}
\end{equation}

If we then substitute the value of $y$ in (\ref{eq:secondorder4}) into (\ref{eq:secondorder3}) we will end up with the more complex model in (\ref{eq:secondorder5}). This represents a form of the generalized Liénard equation \citep{burton1965}, where $h(x, \dot{x})$ is a damping or forcing function that depends on both position and velocity.

\begin{equation}
\dot{y} = -b(0.97\dot{x} + 0.21x - 0.13) - kx
\label{eq:secondorder5}
\end{equation}

While this level of complexity is not necessarily a problem, it results in a much less parsimonious model and probably contains more complexity than necessary. To solve this problem for second-order models, we use Sparse Relaxed Regularized Regression (SR3) \citep{champion-etal2020}. This allows to incorporate constraints on the model, such as placing bounds on coefficient values, forcing terms to be in a particular proportion to one another, or incorporating other forms of physical knowledge about the system. We use SR3-based constraints to enforce the following weak assumptions: (1) $y$ in (\ref{eq:secondorder2}) is always equal $y \mbeq 1.00\dot{x}$, (2) all other potential terms in (\ref{eq:secondorder2}) always equal zero. This essentially constrains the damping function to $-b\dot{x}$ and prohibits additional complexity.

We use a constrained version of the SR3 algorithm that is conceptually analogous to STLSQ but with some important differences, such as the addition of a constraint matrix that makes the solution conditional on the specified constraints. This can be cast as the objective function in (\ref{sr3}), where $\Xi$ is the coefficient matrix to be optimized.

\begin{equation}
\begin{split}
\min\limits_{\Xi, W} \frac{1}{2}||\dot{X} - \Theta(X)\Xi||^{2} + \lambda R(W) + \frac{1}{2\nu}||\Xi - W||^{2}\\
\text{subject to } C\xi = d
\label{sr3}
\end{split}
\end{equation}

The term $\frac{1}{2}||\dot{X} - \Theta(X)\Xi||^{2}$ measures the fit between data and model based on the sum of squared differences, where $\frac{1}{2}$ is a scaling factor that simplifies the derivatives. $R(W)$ is a regularisation function that acts as a prior on sparsity promotion; we specifically use weighted $\ell_{0}$ regularisation, which is a non-convex function that can handle multiple local minima in the optimization landscape. The term $W$ is a proxy variable for $\Xi$ that allows us to decouple model fitting and regularization, such that $\Xi$ can be optimized on the data and subsequently regularized to promote sparsity, with this sequence iterated during the optimization procedure. This improves numerical stability and admits greater flexibility in determining a well-fitting model. The coupling term $\frac{1}{2\nu}||\Xi - W||^{2}$ ensures that $W$ and $\Xi$ remain close and do not substantially diverge from one another. The $\lambda$ parameter weights the regularization function, where $\lambda = \eta^2/2\nu$. This is sparsity-promoting, where $\eta$ is the threshold for the minimum coefficient magnitude in $\Xi$ and $\nu$ determines the closeness of the match between $\Xi$ and $W$.

We use a constrained variant of SR3, where optimizing the objective function in (\ref{sr3}) is subject to a matrix of linear constraints, using the vectorized form $\xi = \text{vec}(\Xi)$. The optimization must meet the condition that $C\xi = d$, where $C$ is a matrix specifying which terms in $\xi$ are subject to constraints, and $d$ is a vector defining the constraint values. In our case, $C\xi$ defines the terms in the first equation $y = \dot{x}$ that are subject to constraints, with $d$ specifying a value of 1.0 for the term $\dot{x}$. This is what allows us to impose the specified constraints on the damping term in second-order differential equations. A proof of the convergence properties of the constrained SR3 algorithm can be found in \citet{champion-etal2020}. In terms of hyperparameters, we use a maximum of 30 iterations to allow for convergence of the optimization algorithm, with $\nu =1$ and a coefficient threshold optimized to maximize $R^2$ model fit from the set $\eta \in \{0.001, 0.01, 0.1\}$.

\subsubsection{Train-test data split}

For the purposes of modelling, we randomly split the data for each articulatory variable into 80\% training and 20\% test sets. The training set is used for library comparison and an initial fit, where the number of discovered terms is allowed to vary between tokens. We then fit the best overall model to the test set, where the model must fit the full library (representing the best model structure) to each trajectory. We also report visualizations of model predictions on a random sample of tokens from the test set. Note that the training set is used to discover a symbolic model, rather than a statistical model, and it is this symbolic model that is fitted to the test set.

\subsubsection{Library selection}

We begin by sensitivity testing the candidate feature library, because the number of terms in a final model can be highly sensitive to the thresholding parameter, especially with single token fits, and this allows us to compare the fits across different feature libraries. As articulatory signals are well approximated as the sum of polynomials, we use a series of polynomial libraries across first ($x$), second ($x, x^2$), third ($x, x^2, x^3$) and fourth ($x, x^2, x^3, x^4$) degrees. We fit each polynomial library to the training data using the model ensembling technique reported below and calculate summary statistics for each library.  This allows us to establish the relative merits of different polynomial libraries and make transparent decisions when two libraries perform very similarly. The selected feature library is then re-fitted to the training data and we report a wider range of summary statistics, which is outlined in more detail below.

The threshold hyperparameter for library comparison was optimized for each articulatory variable from the set $\{0.001, 0.01, 0.1\}$, with the final threshold value based on the highest $R^2$ value. In some cases, lower thresholds did not converge due to an ill-conditioned or stiff model that resulted in numerically unstable predictions. In other cases, higher polynomial libraries performed worse than lower polynomial libraries. This can appear surprising, because adding an additional polynomial term to a well-performing model should not harm performance. This phenomenon arises, however, because a greater number of terms often reconfigures the model in a way that changes the magnitude of each coefficient term. In other words, a quadratic model is not necessarily the linear model with its original coefficients plus a quadratic term, but can sometimes be a fundamentally different model in terms of the relationship between coefficients. As a result, even a very small threshold can eliminate key terms in such models, or result in higher polynomials fitting to noise in the signal. We address this by sensitivity testing threshold values as above and selecting feature libraries that exhibit stability across the data set. In cases where additional complexity provides only minor performance improvements, we subsequently explore whether greater complexity simply improves the quantitative fit of models, or reveals fundamentally different qualitative dynamics of the system.

\subsubsection{Model ensembling}

In order to discover models across large data sets we use a model ensembling technique, whereby models are fitted to each token separately and then an ensemble model is derived from this set of models. While a single SINDy model can be fitted using multiple trajectories, simulations show that this is generally only effective for speech gestures if the target or equilibrium position is the same or very similar across tokens. In such a case, a single model and single set of coefficient values will be returned for the whole data set. However, when key parameters vary -- such as stiffness, damping or target -- then the resulting SINDy model will contain values for these parameters that are the average best fit to all of the tokens. As we seek good symbolic models and accurate parameters, we instead construct an ensemble model from models fitted to individual trajectories. As the model discovery procedure is repeated for each token, each trajectory could theoretically be fitted with different numbers of terms, especially in the case of larger feature libraries. We leverage this fact in order to obtain distributions on the number of terms in each model across the data set, which allows us to then arrive at an ensemble model based on the majority model structure. Once this final model structure is determined, we then fit it to the test data, forcing the same structure on each test trajectory.

\subsubsection{Generating predictions}

Once we have discovered a model for each token, we then use this model to make a prediction. Predictions are generated by taking the discovered model, the discovered parameter coefficients, and a set of initial conditions from the data, which comprise the initial position and velocity value in each empirical trajectory. These initial conditions are then used to solve the discovered model forwards in time, determined by the discovered coefficient values. In an ideal scenario, this should generate position and velocity trajectories that are identical to the original data. In practice, however, the prediction is only as good as the discovered model, which allows us to use the prediction as an estimate of model fit. We quantify the fit using by-trajectory $R^2$ scores. All $R^2$ values are variance weighted, meaning that the $R^2$ for model fit is an average over position and velocity, weighted by the variance of each signal, which provides a more informative assessment of model performance given that position and velocity are on different numerical scales. In conventional regression analysis, $R^2$ values are bounded between [0,1], but the lower bound is a consequence of allowing either the intercept or the slope to vary, which is the aim of regression analysis. With a constrained intercept, however, $R^2$ is negative when the prediction is worse than simply fitting a horizontal line through the data \citep{chicco-etal2021}. We fix the intercept for each model prediction as the initial conditions from empirical data, so $R^2$ will be negative when the model prediction is worse than a horizontal line through the data.

\section{Discovering new models from data}
\label{sec:analysis}

\subsection{First-order models}
\label{subsec:first_order}

\subsubsection{Library comparison}

Table \ref{table:sentences-firstorder-poly} shows the results for library comparison on first-order models. The optimal threshold for the models was $\lambda$ = 0.001 (LA), 0.1 (TT), 0.01 (TD), 0.001 (TR). The first-degree library performs poorly, with $R^2$ values across the four articulatory variables of \{0.40, 0.56, 0.53, 0.51\}. Libraries with 2--4 polynomials perform well with $R^2$ between 0.96--0.99. Overall, while the second-degree library performs well on average, it contains some negative $R^2$ values for LA, TT and TD. By contrast, the lowest score for any articulatory variables in the third-degree library is $R^2$ = 0.63 (LA). For this reason, we select the third-degree library containing $x, x^2, x^3$, which we more thoroughly evaluate in the following section.

\begin{table}[h]
\centering
\begin{tabular}{rrrrrr}
\hline
Number of polynomials  & LA & TT & TD & TR\\
\hline
1		& 0.40 (0.57)   &  0.56 (0.17)	& 0.53 (0.24)	& 0.51 (0.29)  \\
2		&  0.96 (0.05)  & 0.96 (0.05)	& 0.96 (0.05)	& 0.96 (0.03)  \\
3		&  0.96 (0.04) & 0.97 (0.04)	& 0.96 (0.03)	& 0.96 (0.03)\\
4	 	& 0.98 (0.02)   & 0.98 (0.02)	& 0.99 (0.02)	& 0.99 (0.02) \\
\hline
\end{tabular}
\caption{Comparison of different polynomial libraries in first-order models across articulatory variables (training set). The model scores are $R^2$ mean (standard deviations).}
\label{table:sentences-firstorder-poly}
\end{table}

\subsubsection{Results}

Table \ref{table:sentences-firstorder-model} shows summary statistics for the first-order models with third-degree polynomial library fitted to the training data. All four terms (a constant, $x, x^2, x^3$) were found for the majority of models (LA = 99.78\%, TT = 64.67\%, TD = 93.68\%, TR = 99.64\%), but some models omitted the cubic term (LA = 0.22\%, TT = 35.23\%, TD = 6.28\%, TR = 0.36\%) and a small percentage comprised only linear terms (TT = 0.1\%, TD = 0.04\%)

\begin{table}[h]
\centering
\begin{tabular}{rrrrr}
\hline
Training set	& LA & TT & TD & TR \\
\hline
mean($R^2$)		& 0.96	& 0.97	& 0.96	& 0.96	\\
$\sigma$($R^2$)	& 0.04	& 0.02	& 0.03	& 0.03	\\
min($R^2$)		& 0.63	& 0.66	& 0.63	& 0.45	\\
max($R^2$)		& 1.00	& 1.00	& 1.00	& 1.00	\\
\hline
\end{tabular}
\begin{tabular}{rrrrr}
\hline
Test set	& LA & TT & TD & TR \\
\hline
mean($R^2$)		& 0.95	& 0.95	& 0.95	& 0.95	\\
$\sigma$($R^2$)	& 0.01	& 0.02	& 0.01	& 0.02	\\
min($R^2$)		& 0.77	& 0.78	& 0.84	& 0.84	\\
max($R^2$)		& 0.99	& 0.99	& 0.99	& 0.99	\\
\hline
\end{tabular}
\caption{Model fit statistics for first-order models with polynomials up to third-degree. All values rounded to 2 decimal places.}
\label{table:sentences-firstorder-model}
\end{table}

All models perform fairly well, with mean $R^2$ = 0.96 or above and minimum $R^2$ of 0.45--0.66. The test data was then fitted using the same algorithm, but all trajectories were forced to have linear, quadratic and cubic terms. The test data show comparable performance, with mean $R^2$ = 0.95 for each articulatory variable and higher minimum $R^2$ values in each case, ranging from 0.77--0.84. The higher performance in the test data is because every token is forced to contain all terms in the library (i.e. $\lambda = 0$).

\begin{figure}[h]
    \centering
    \begin{subfigure}[b]{0.49\textwidth}
        \includegraphics[width=\textwidth]{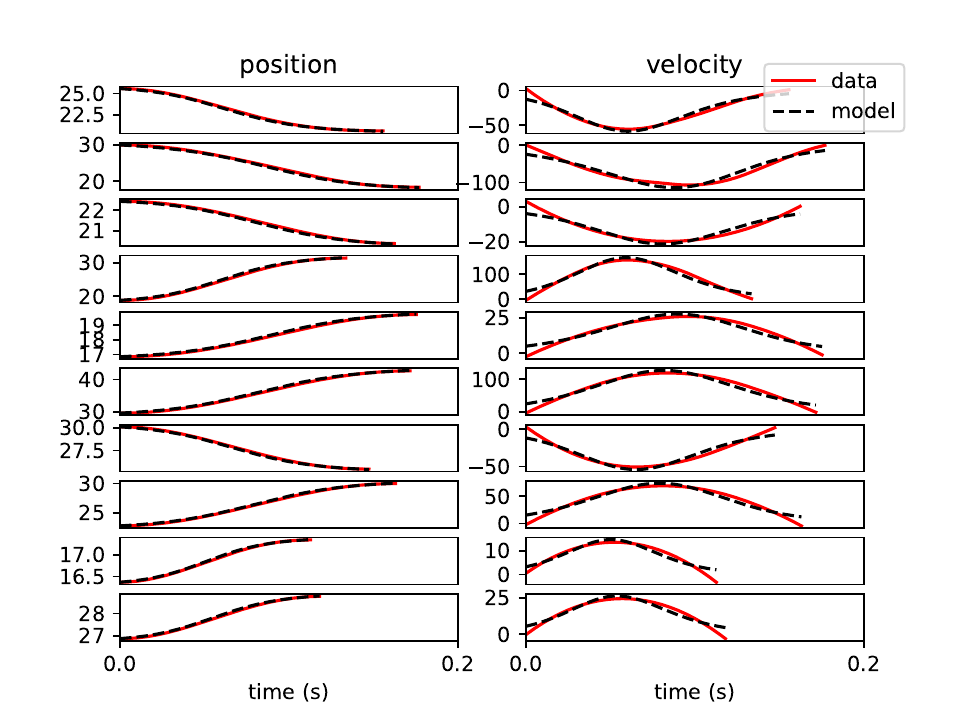}
        \caption{LA (Lip Aperture)}
    \end{subfigure}
    \begin{subfigure}[b]{0.49\textwidth}
        \includegraphics[width=\textwidth]{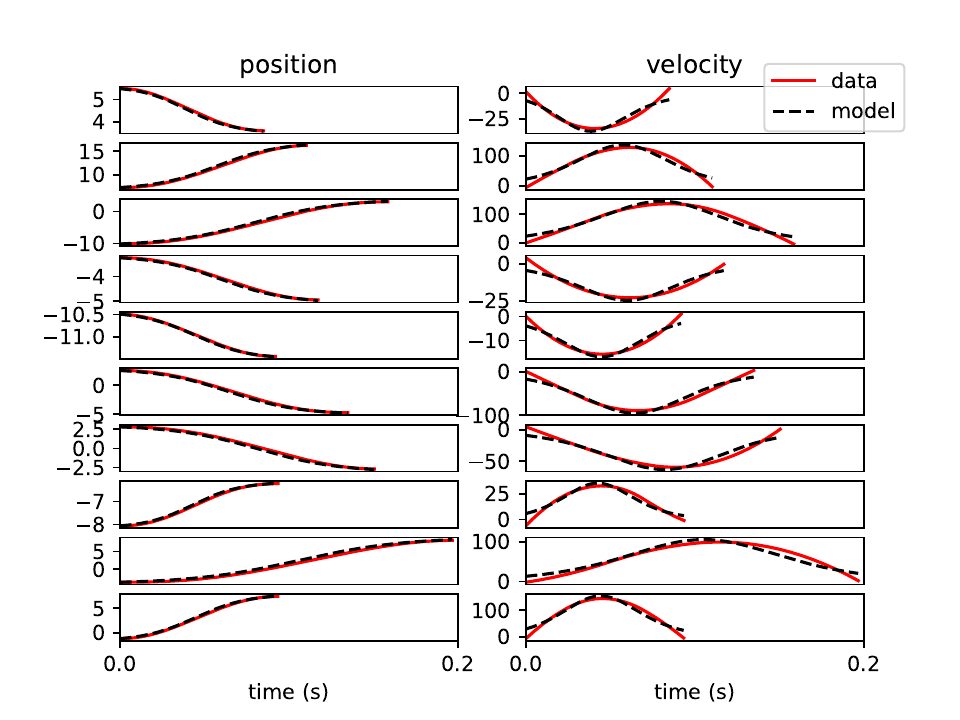}
        \caption{TT (Tongue Tip)}
    \end{subfigure} 
    
    \begin{subfigure}[b]{0.49\textwidth}
        \includegraphics[width=\textwidth]{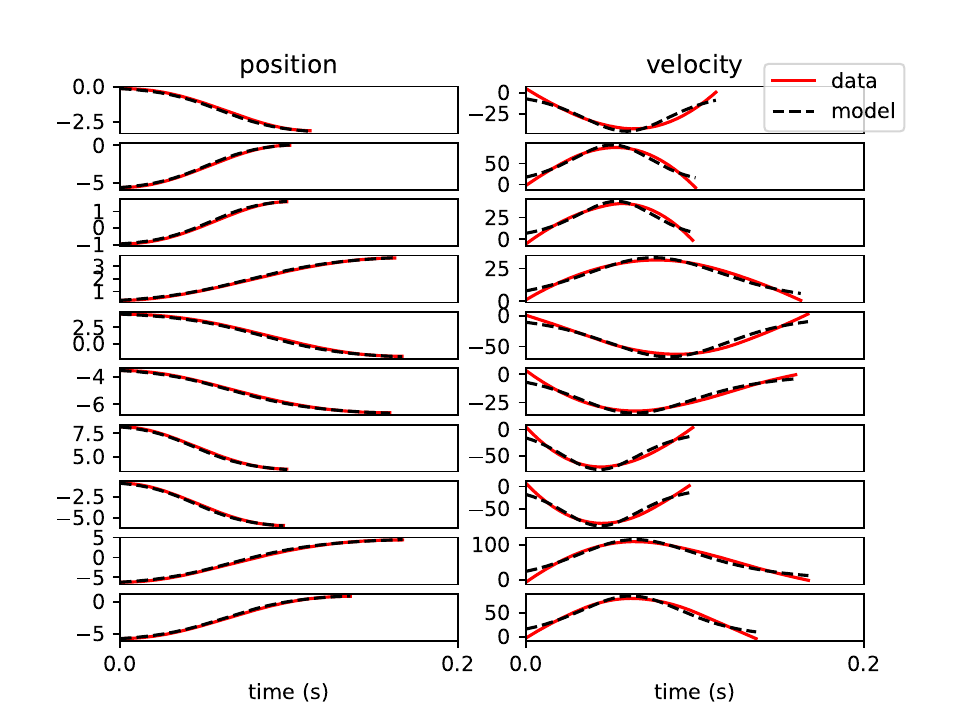}
        \caption{TD (Tongue Dorsum)}
    \end{subfigure}
    \begin{subfigure}[b]{0.49\textwidth}
        \includegraphics[width=\textwidth]{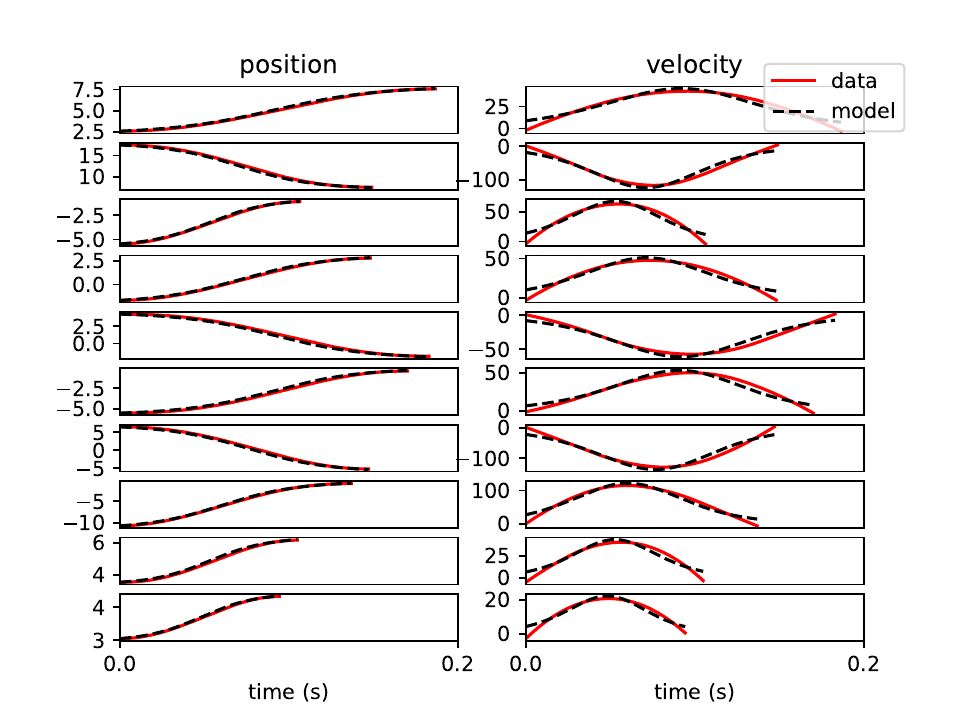}
        \caption{TR (Tongue Root)}
    \end{subfigure}
\caption{10 randomly sampled trajectories for each articulatory variable showing the fit between data and model predictions for first-order models with a third-degree polynomial library on the test data.}
\label{plot:sentences-firstorder}
\end{figure}

Figure \ref{plot:sentences-firstorder} shows 10 randomly-sampled position and velocity trajectories for each variable, with a comparison of data and model prediction from the test data. The quantitative fit is good, but with some errors in the intercept and also errors in the fit along the curves. As a result of minor errors in the position data, the resulting velocity signals show bigger errors. This is a consequence of a limited degree of flexibility in the possible shape of the velocity curve, due to the lack of an acceleration term governing change in velocity.

\subsubsection{Summary}

The discovered model takes the form of the symbolic equation in (\ref{eq:sentences-first-order}), where $a$ is a constant and $b, c, d$ are the coefficients of $x, x^2, x^3$. This makes the discovered model a cubic equation, where velocity is dependent on a constant, as well as the current position multiplied by a coefficient, the square of the current position multiplied by a coefficient, and the cube of the current position multiplied by a coefficient. The presence of $x^2, x^3$ makes it a nonlinear model.

\begin{equation}
\dot{x} = a - bx + cx^2 - dx^3
\label{eq:sentences-first-order}
\end{equation}

In summary, a first-order nonlinear model appears to be a reasonable quantitative fit across four different articulatory variables.

\subsection{Second-order models}
\label{subsec:second_order}

\subsubsection{Library comparison}

Table \ref{table:reps-secondorder-poly} shows the library comparison on second-order models. In this instance, the first, second and third-degree libraries perform near-identically in terms of summary statistics, with mean $R^2$ = 0.98--0.99, although the first-degree models have slightly lower standard deviations. The optimal threshold for all models was $\eta = 0.001$. The fourth degree library performs poorly with a very small threshold of $\eta = 0.001$; if we relax the threshold to zero then the fourth-degree library performs marginally better than all other libraries, but there is clearly no need for a model of this level of complexity. In addition to this, all models above second-degree have some negative $R^2$ fits, suggesting that the greater complexity forces some important coefficients to values smaller than the threshold. Overall, there appears to be little benefit in higher polynomial libraries based on this comparison. As result, we select the first degree library, but evaluate the impacts of any additional complexity in Section \ref{sec:models_secondorder_cubic}.

\begin{table}[ht]
\centering
\begin{tabular}{rrrrr}
\hline
Number of polynomials & LA & TT & TD & TR \\
\hline
1	& 0.99 (0.02)		&  0.98 (0.03)		& 0.98 (0.02)		& 0.99 (0.02)    \\
2      &  0.99 (0.01)		& 0.99 (0.01)		& 0.99 (0.01)		& 0.99 (0.01)  \\
3	&  0.99 (0.05)		& 0.98 (0.08)		& 0.99 (0.07)		& 0.99 (0.03) \\
4	& $-$1.14 (6.26)	& $-$3.24 (14.26)	& $-$3.05 (19.57)	& $-$2.00 (10.73)  \\
\hline
\end{tabular}
\caption{Comparison of different polynomial libraries in second-order models across articulatory variables (training set). The model scores are $R^2$ mean (standard deviations).}
\label{table:reps-secondorder-poly}
\end{table}

\subsubsection{Results}

Table \ref{table:sentences-secondorder-model} shows summary statistics for the second-order models with first-degree polynomial library. In the training data, 100\% of models contain 3 terms, with mean $R^2 > 0.98$ in all cases. The test data shows comparable performance, with minimum $R^2$ values of \{0.77, 0.72, 0.66, 0.76\}.

\begin{table}[ht]
\centering
\begin{tabular}{rrrrr}
\hline
Training set		& LA		& TT		& TD		& TR \\
\hline
mean($R^2$)		& 0.99	& 0.98	& 0.98	& 0.99	\\
$\sigma$($R^2$)	& 0.02	& 0.03	& 0.02	& 0.02	\\
min($R^2$)		& 0.75	& 0.70	& 0.64	& 0.70	\\
max($R^2$)		& 1.00	& 1.00	& 1.00	& 1.00	\\
\hline
\end{tabular}
\begin{tabular}{rrrrr}
\hline
Test set			& LA		& TT		& TD		& TR \\
\hline
mean($R^2$)		& 0.99	& 0.98	& 0.99	& 0.98	\\
$\sigma$($R^2$)	& 0.02	& 0.03	& 0.02	& 0.02	\\
min($R^2$)		& 0.77	& 0.72	& 0.66	& 0.76	\\
max($R^2$)		& 1.00	& 1.00	& 1.00	& 1.00	\\
\hline
\end{tabular}
\caption{Model fit statistics for second-order models with polynomials up to first-degree. All values rounded to 2 decimal places.}
\label{table:sentences-secondorder-model}
\end{table}

\begin{figure}[]
    \centering
    \begin{subfigure}[b]{0.49\textwidth}
        \includegraphics[width=\textwidth]{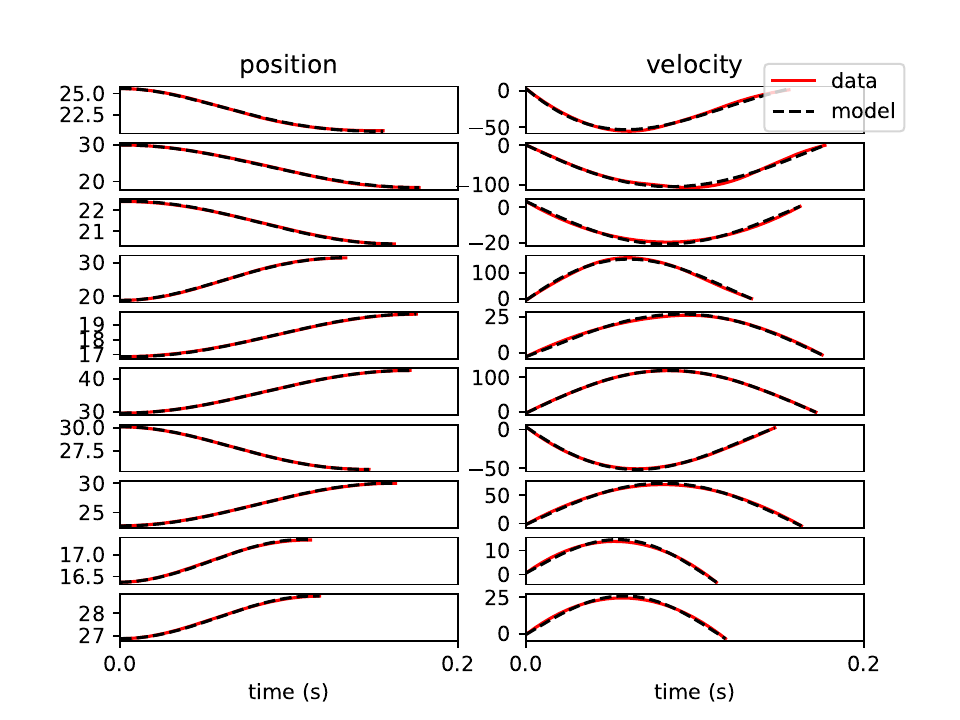}
        \caption{LA (Lip Aperture)}
    \end{subfigure}
    \begin{subfigure}[b]{0.49\textwidth}
        \includegraphics[width=\textwidth]{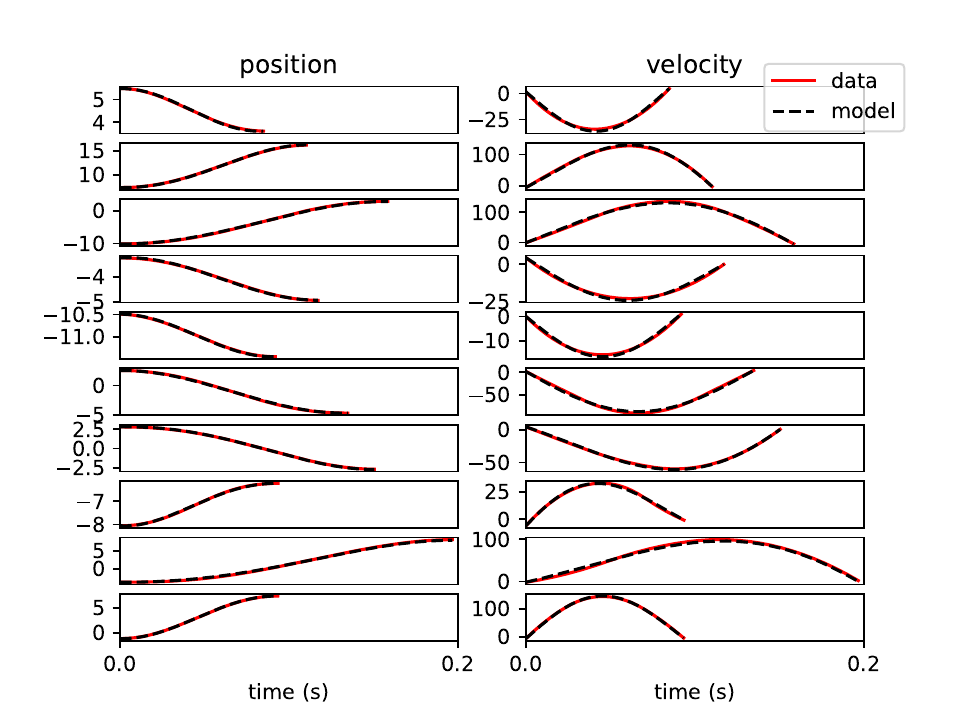}
        \caption{TT (Tongue Tip)}
    \end{subfigure} 
    
    \begin{subfigure}[b]{0.49\textwidth}
        \includegraphics[width=\textwidth]{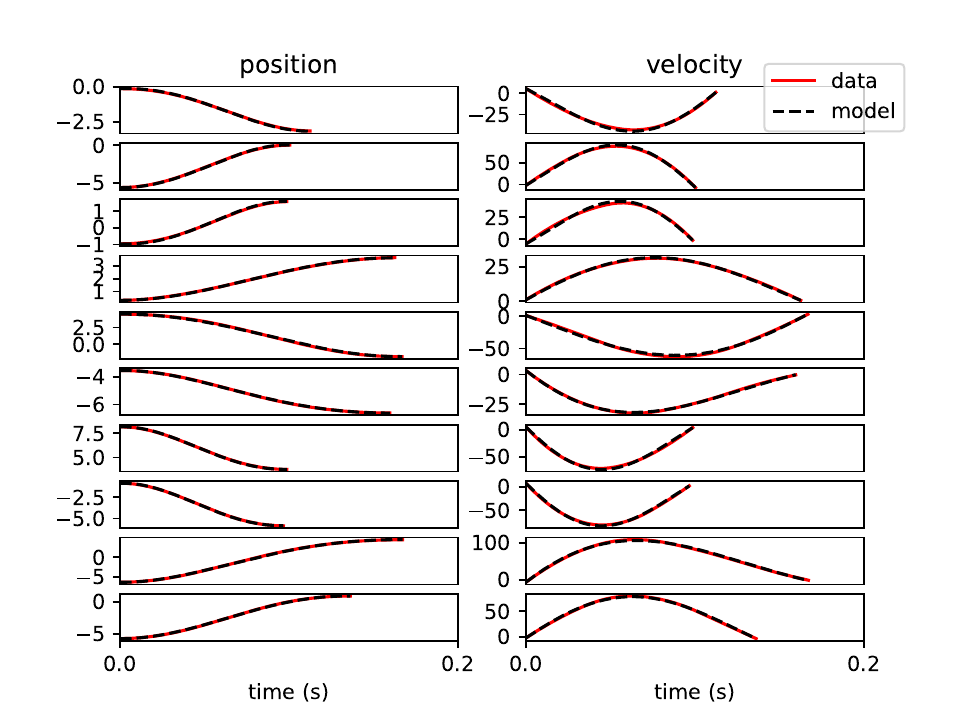}
        \caption{TD (Tongue Dorsum)}
    \end{subfigure}
    \begin{subfigure}[b]{0.49\textwidth}
        \includegraphics[width=\textwidth]{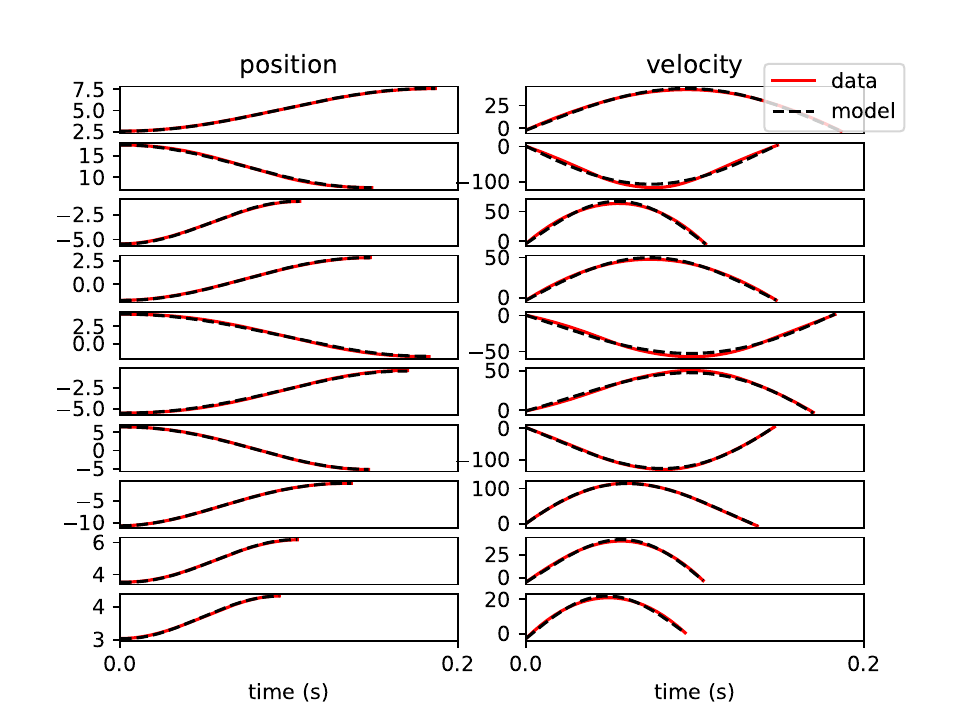}
        \caption{TR (Tongue Root)}
    \end{subfigure}
\caption{10 randomly sampled trajectories for each articulatory variable showing the fit between data and model predictions for second-order models with a first-degree polynomial library on the test data.}
\label{plot:sentences-secondorder}
\end{figure}

Figure \ref{plot:sentences-secondorder} shows 10 randomly-sampled position and velocity trajectories for each variable, with a comparison of data and model prediction from the test data. The quantitative fit is excellent, with near-perfect fits for all trajectories. Notably, the fits look substantially better than those in Figure \ref{plot:sentences-firstorder}, suggesting that the second-order model is superior in quantitative fitting accuracy.

\subsubsection{Summary}

The discovered model takes the form of the symbolic equation in (\ref{eq:sentence-second-order}).

\begin{equation}
\ddot{x} = kT - kx - b\dot{x}
\label{eq:sentence-second-order}
\end{equation}

Note that SINDy models sometimes discover a constant term, but the analysis in Section \ref{subsec:sim_example} shows that this is often the term $kT$, such as that $kT -kx = -k(x-T)$. As such, this model takes the form the harmonic oscillator in (\ref{eq:sentence-second-order2}), which is a standard task dynamic model \citep{saltzman-munhall1989}.

\begin{equation}
\ddot{x} = -b\dot{x} -  k(x - T)
\label{eq:sentence-second-order2}
\end{equation}

In summary, a second-order linear model is a very good fit to data across four different articulatory variables. The mean accuracy of the fit is $R^2 = 0.98$ and above in all cases, with no trajectories being scored less than $R^2$ = 0.64 in either the training or test data sets. In Section \ref{sec:models}, we conduct further interpretation of these terms and explore the implications of this model.

\subsection{Interim summary}

Two models fit the data very well: a first-order nonlinear model with quadratic and cubic terms, and a second-order linear model. The performance of both models is quantitatively similar, but this may be a consequence of the mean scores for both models being relatively close to ceiling. An inspection of random trajectories plotted from each model reveals that while both models fit the data well, the second-order fits are more accurate, likely a consequence of the acceleration term in the second-order models. The following sections take the two broad classes of models discovered here and explore them in greater detail. Specifically, we interpret the meaning of each model's terms in light of known systems and explore the model space via computational simulations.

\section{Exploring the discovered models}
\label{sec:models}

\subsection{Overview}

So far, we have two well-fitting models, but at this point we must go beyond treating these simply as effective fits to data and understand the ways in which they govern the dynamical laws of speech. In this section, we take two models from Section \ref{sec:analysis}  -- a first-order model and a second-order model -- and explore them deeper, focusing on how to interpret the different terms in the equations, as well as what predictions and assumptions they make about speech gestures. In doing so, we also explore the effects of adding complexity to the second-order model.

The analysis for each section proceeds as follows. We first plot representative examples of the data against model predictions, followed by plotting the qualitative dynamics of the system in the form of phase portraits and Hooke portraits \citep{beek-beek1988, mottet-bootsma1999}. This is an essential step in moving beyond model assessment via simple data fitting, because a fundamental characteristic of dynamical models of skilled movement is how they specify the relationship between position, velocity and acceleration. We then explore the equations analytically, deriving algebraic properties that expose similarities to other well-understood systems, before exploring the dynamics of the relevant terms using computational simulations. We note that a comprehensive investigation of every aspect of each model and its numerical parameterization is beyond the scope of the current article; instead, we here focus on elucidating fundamental aspects of each system.

\subsection{First-order model}
\label{sec:models_firstorder}

\begin{figure}[h]
\centering
\includegraphics[scale=0.6]{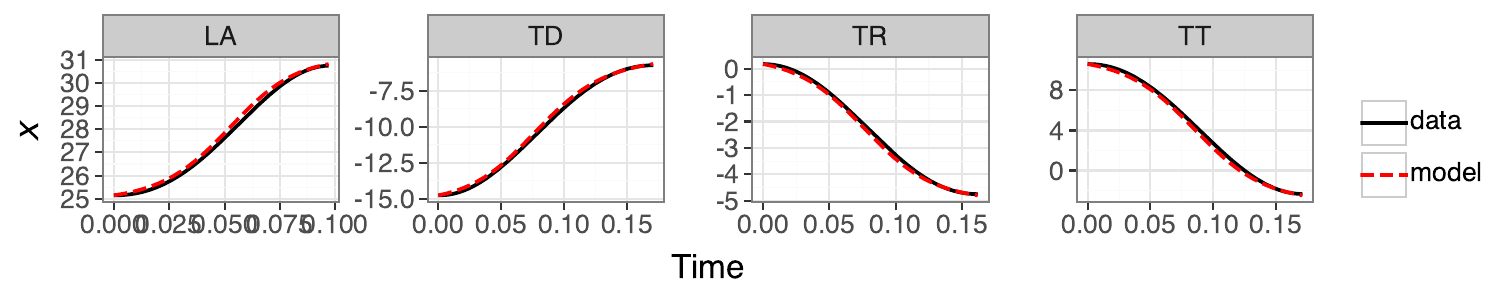}
\includegraphics[scale=0.6]{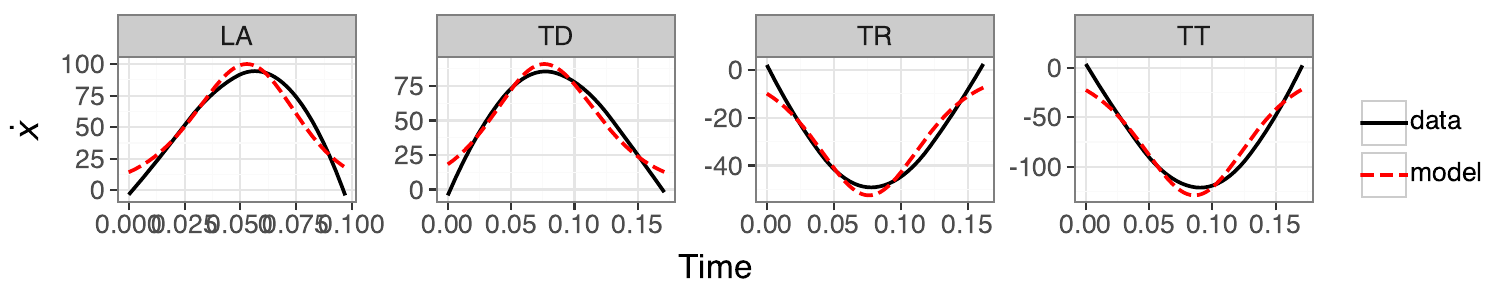}
\includegraphics[scale=0.6]{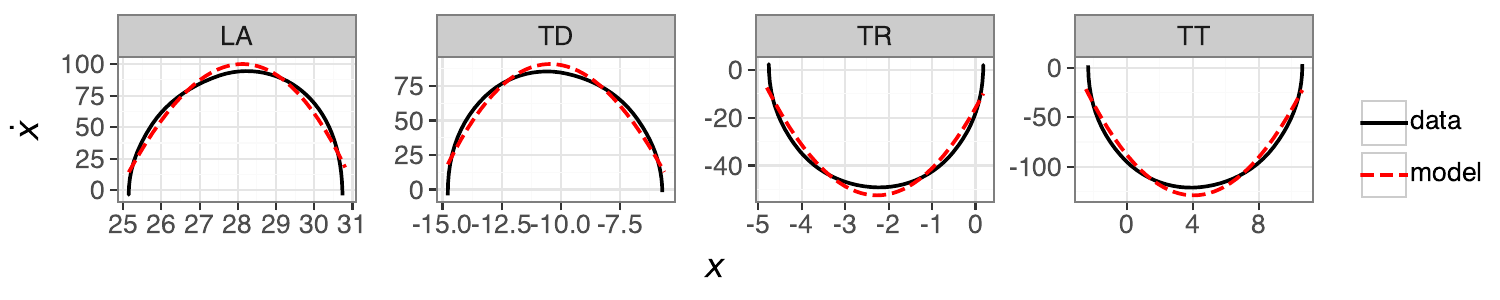}
\includegraphics[scale=0.6]{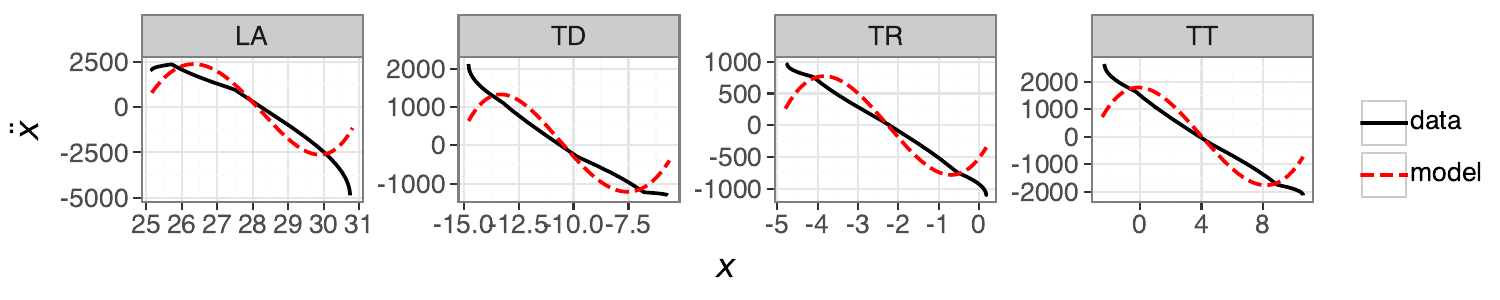}
\caption{First-order model predictions showing time-varying position (row 1), time-varying velocity (row 2), phase portrait (row 3), Hooke portrait (row 4). The $R^2$ scores for each model are LA = 0.96, TD = 0.95, TR = 0.96, TT = 0.95.}
\label{plot:firstorder_phase}
\end{figure}

Figure \ref{plot:firstorder_phase} shows representative examples of the first-order model predictions, with position and velocity trajectories, as well as a phase portrait (position $\sim$ velocity) and Hooke portrait (position $\sim$ acceleration). While the position data are predicted with good accuracy, the higher derivatives and phase portraits show the inadequacy of a first-order model. The velocity trajectory and phase portrait show how the velocity estimates at signal edges are systematically incorrect, as well as showing small but systematic errors in peak velocity and time-to-peak velocity. The Hooke portrait particularly highlights the poor predictions of the first-order model, where uniform nonlinearity is predicted despite quasi-linearity in the empirical data.

\begin{figure}[h]
\centering
\includegraphics[scale=0.6]{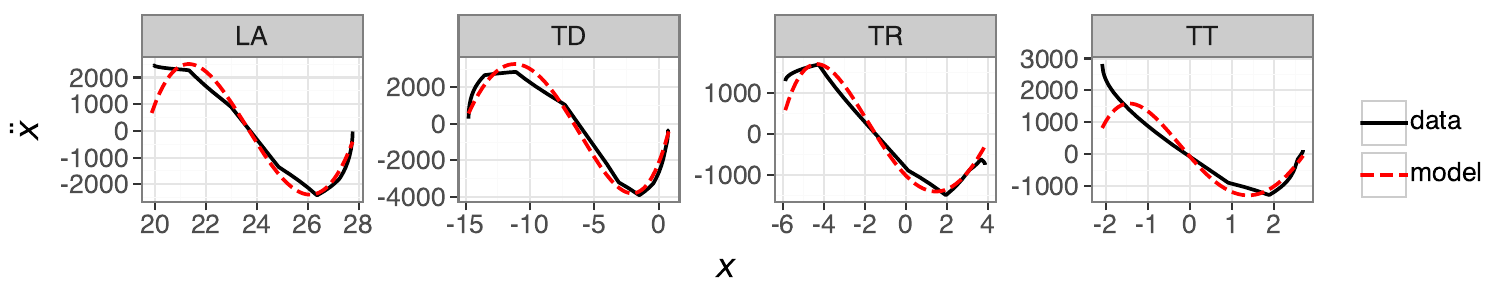}
\includegraphics[scale=0.6]{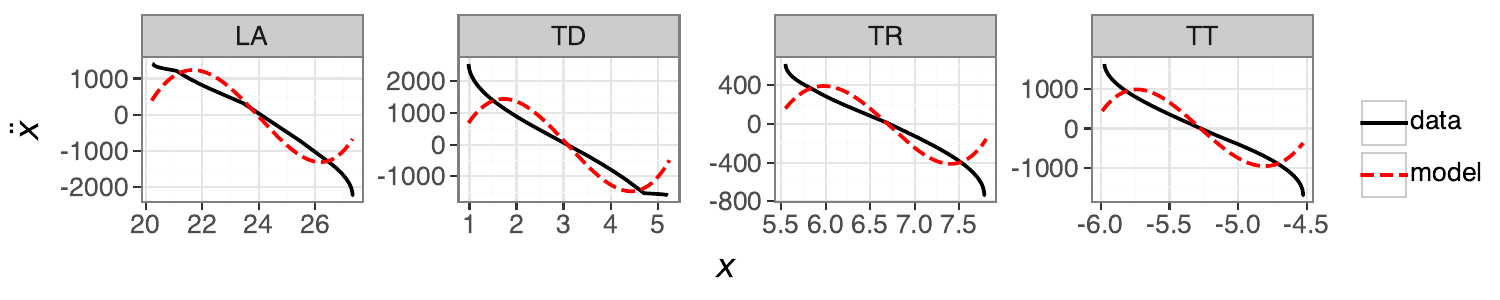}
\caption{Hooke portraits for the top-scoring (top row) and median-scoring (bottom row) first-order models for each articulatory variable.}
\label{plot:firstorder_hooke}
\end{figure}

The above suggests that $R^2$ scoring of position and velocity leads to misleading conclusions about the true capacity of the first-order model. To explore this further, Figure \ref{plot:firstorder_hooke} shows Hooke portraits of data and predictions for the best-scoring (top row) and median-scoring (bottom row) first-order nonlinear models. The best fitting model is a case with extensive nonlinearity between position and acceleration, especially for the TD articulatory variable. However, similarly nonlinear predictions are also made for the median-scoring models, despite the data showing a quasi-linear relation in these cases. This is a consequence of the first-order model containing no information about the system's higher derivatives, such as acceleration; this makes it impossible to accurately model change in velocity, which is particularly evident at movement onsets. 

In conclusion, this analysis shows that a first-order nonlinear equation is not an appropriate model of articulatory control. While the SINDy analysis revealed a good fit to empirical position and velocity trajectories, this did not take into account the \textit{nature} of the mismatches between model and data. In this instance, the phase portrait and Hooke portrait point towards a fundamental issue with the model, rather than minor errors in quantitative fit. These findings suggest that this model is insufficient for capturing the dynamical characteristics of articulatory movements. In addition, the first-order model contains a greater number of parameters (with quadratic and cubic terms), which is inevitable given the lack of terms for controlling higher derivatives. This shows that the first-order model's complexity is not warranted due to its theoretical inadequacy. As a consequence, we do not consider this model any further and move on to the second-order linear model.

\subsection{Second-order linear model}
\label{sec:models_secondorder}

\begin{figure}[h]
\centering
\includegraphics[scale=0.6]{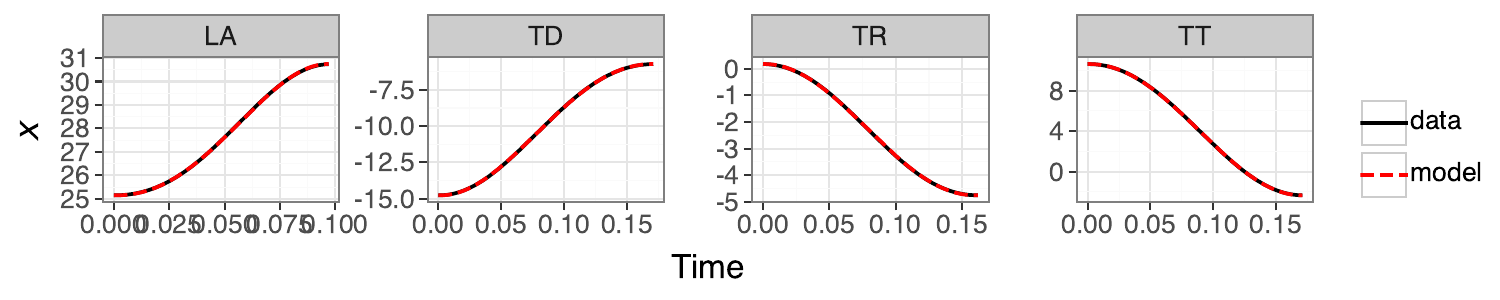}
\includegraphics[scale=0.6]{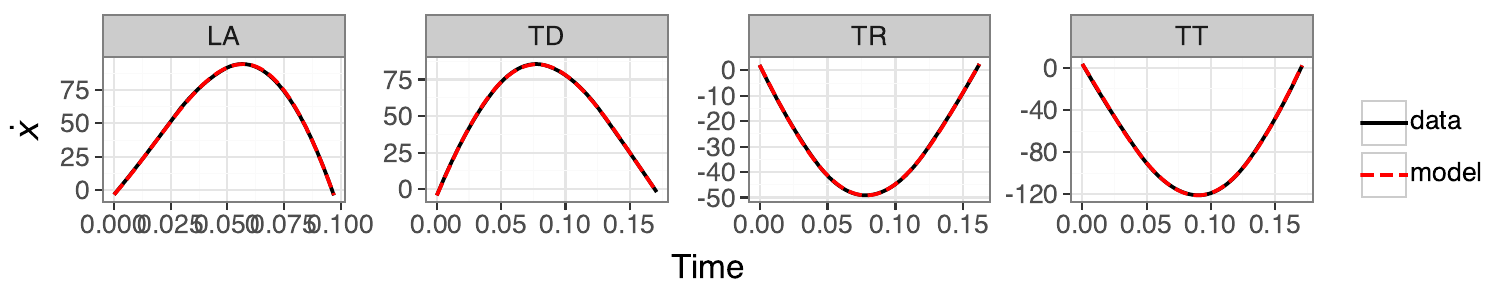}
\includegraphics[scale=0.6]{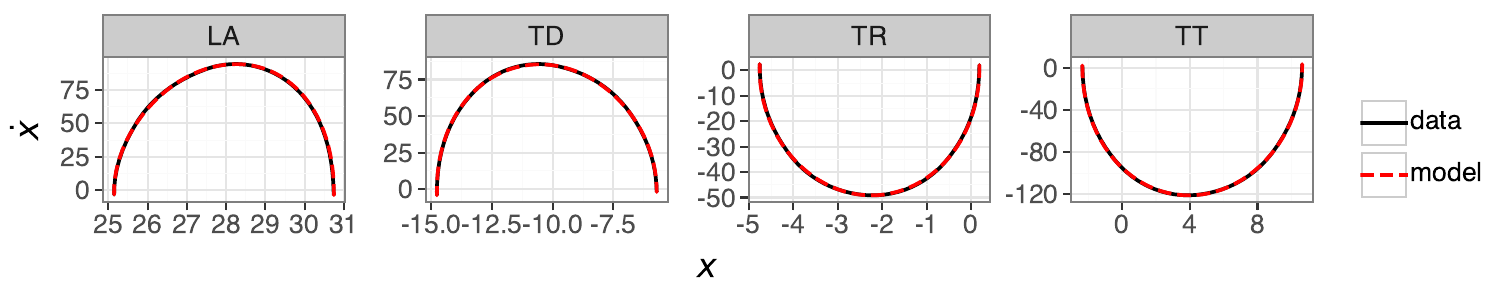}
\includegraphics[scale=0.6]{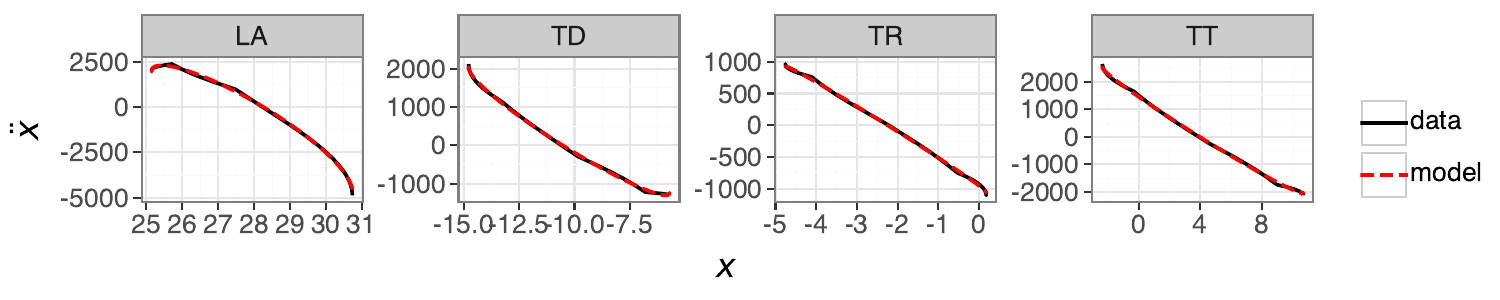}
\caption{Linear second-order model predictions showing time-varying position (row 1), time-varying velocity (row 2), phase portrait (row 3), Hooke portrait (row 4). The SINDy scores for each model are LA = 0.999, TD = 0.999, TR = 0.999, TT = 0.999}
\label{plot:secondorder_phase}
\end{figure}

Figure \ref{plot:secondorder_phase} shows data and predictions for the best-fitting second-order model for each articulatory variable. In contrast to the first-order model in Section \ref{sec:models_firstorder}, the second-order model is an excellent fit across position and velocity, the corresponding phase portrait, and the Hooke portrait. Notably, the Hooke portraits show quasi-linearity, suggesting that the dynamics are well approximated by harmonic motion \citep{beek-beek1988}. This is a key signature of a linear oscillator, in contrast to the anharmonicity that would suggest nonlinear dynamics.

We note that Figure \ref{plot:secondorder_phase} captures cases that correspond to highly linear models. The remainder of this section focuses on these cases and their interpretation, but in Section \ref{sec:models_secondorder_cubic} we explore cases where the linear model fits worse and consider whether additional complexity is warranted. For now, we turn to the second-order model that was discovered in Section \ref{sec:discovering_models}. This equation took the form in (\ref{eq:second-order-form}):

\begin{equation}
\ddot{x} = - b\dot{x} -k(x-T)
\label{eq:second-order-form}
\end{equation}

\begin{figure}[h]
\centering
\includegraphics[scale=0.6]{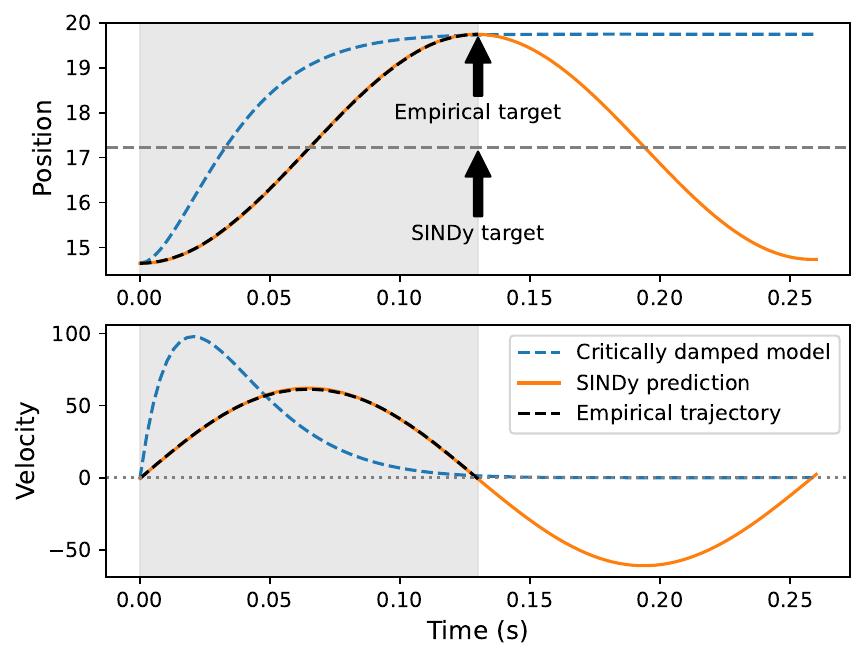}
\caption{Dynamics of the second-order linear model with SINDy-discovered parameters. The dashed black line is the empirical trajectory with the best fitting model, the orange line is the SINDy model prediction ($T=17.22, k = 592, b = 0.264, R^2 = 0.999$), the dashed blue line represents the coefficients needed to reach the empirical target under the assumption of critical damping ($T = 19.75, k=2500, b = 2 \sqrt{k}$). The grey shaded area corresponds to the duration of the empirical trajectory, with the subsequent unshaded region showing the effects of continuing the simulation beyond this duration. The label `SINDy target' corresponds to the SINDy-discovered value of $T$, whereas the label `empirical target' is the empirical position value corresponding to the final velocity minimum.}
\label{plot:secondorder_sindy}
\end{figure}

This corresponds to the form of a linear harmonic oscillator, but it is clear that the system is not critically damped, as a critically damped version of (\ref{eq:second-order-form}) is unable to fit the symmetrical empirical velocity profiles as accurately as seen in Section \ref{subsec:second_order} \citep{sorensen-gafos2016}. How, then, does the system achieve its equilibrium position? Figure \ref{plot:secondorder_sindy} shows the trajectory with the best fitting SINDy model ($R^2 = 0.999$), which is a Tongue Root trajectory produced by speaker JW45. The grey shaded area shows the empirical duration of the trajectory, with the dashed black line showing the empirical data. The orange line shows SINDy predictions, which are near-perfect fits to the data, but with an important characteristic: this is only true during the time period that corresponds to the empirical trajectory. If we continue the simulation beyond the trajectory's original duration the system begins to oscillate.

To unpack this further, the dashed blue trajectory shows the trajectory that a critically damped oscillator would need to take to reach the empirical target. This does reach the target, but with a very early time-to-peak velocity, demonstrating the poor empirical fit of a critically damped model. In contrast, the behaviour that the SINDy parameters capture is as follows. It drives the system towards an equilibrium value $T = 17.22$ , which is the SINDy discovered coefficient for $T$. This is below the empirical target $T = 19.75$ and is clearly not the true target in the sense of the gestural system (i.e. the position value at the final velocity zero-crossing). To disambiguate, we henceforth refer to the actual empirical target at the velocity zero-crossing as $T$ and SINDy's discovered virtual target as $T_{v}$ (the term `virtual target' is used entirely to refer to the SINDy discovered target and we make no claims about its theoretical status at this stage). Importantly, the SINDy discovered trajectory shows oscillatory behaviour if extended beyond the empirical duration, as shown by the period after the grey shaded area (i.e. $t > 0.13$).

How are we to relate the SINDy target $T_{v} = 17.22$ to the empirical target $T = 19.75$? The horizontal dashed line in Figure \ref{plot:secondorder_sindy} shows the value of $T_{v}$. We can see that the orange line reaches this target twice: once before empirical $T$ and once after empirical $T$, which is only visible if we extend the simulation beyond the empirical duration of the fitted trajectory. The oscillation is due to the lack of damping $b \approx 0$, but there is a strong intrinsic relationship between $T$ and $T_{v}$, such that  $T_{v}$ can be easily derived from (\ref{eq:t_virtual}), which is equivalent to saying that the virtual target is half the distance between the initial condition and $T$. We can then solve for $T$ as in (\ref{eq:t_actual}).

\begin{equation}
T_{v} = x_{0} + \frac{T-x_{0}}{2}
\label{eq:t_virtual}
\end{equation}

\begin{equation}
T= 2T_{v} - x_{0}
\label{eq:t_actual}
\end{equation}

If we substitute (\ref{eq:t_actual}) into our second-order harmonic oscillator model and rearrange terms then we get (\ref{eq:secondorder_final}). This suggests an alternative hypothesis, where the model contains $\frac{k}{2}(T + x_{0})$. This new equation still requires tuning of $b, k$ to avoid the system producing oscillations when $b < 2 \sqrt k$, which we address below, but it allows us to formulate a model where $T$ captures the empirically-observed target of the system. The only new parameter in the model is $x_{0}$ (initial position), which any dynamical system necessarily already has access to in calculating a trajectory, so it is still an autonomous system. The only modification here is the use of a more complex constant term $\frac{k}{2}(T + x_{0})$.

\begin{equation}
\ddot{x} = -b\dot{x} - kx + \frac{k}{2}(T + x_{0})
\label{eq:secondorder_final}
\end{equation}

The new equation captures cases where the velocity trajectory is very close to a half-cycle sine wave, representing an undamped or minimally damped oscillator where $b \approx 0$. For example, if we use absolute-valued SINDy-discovered parameters for $T, x_{0}$ based on Equation (\ref{eq:t_actual}) then $|T|$ is strongly correlated with the positional value of $|x|$ at the final velocity minimum during LA  ($r = 0.94$) and TR ($r = 0.91$), but only moderately correlated for TT ($r = 0.71$) and TD ($r = 0.82$). There appears to be no discernible difference between opening/closing gestures, with only minimally higher correlations for closing gestures (e.g. largest difference is LA closing $r$ = 0.94 vs. LA opening $r$ = 0.92). The differences between articulatory variables could be a consequence of potential differences in velocity segmentation quality. For example, correlations under the simpler formulation $\ddot{x} = -b\dot{x} - k(x-T)$ are LA = $-0.79$, TT =  $-0.72$, TD = $-0.83$, TR = $-0.91$, suggesting that only LA benefits from the formulation in (\ref{eq:secondorder_final}). The other articulatory variables show greater variability in this respect, although the above example shows a case where TR velocity is highly symmetrical, suggesting that it improves model fit for some tokens, but is not otherwise detrimental. Despite this, we note that the mean $b$ values differ only slightly between articulatory variables, so this model's improved performance for LA could be a consequence of data processing (e.g. one-dimensional lip aperture vs. compression of horizontal/vertical movements into a single dimension) or other properties of lip movements (e.g. reduced coarticulation from neighbouring lingual movements).

\begin{figure}[ht]
\centering
\includegraphics[scale=0.5]{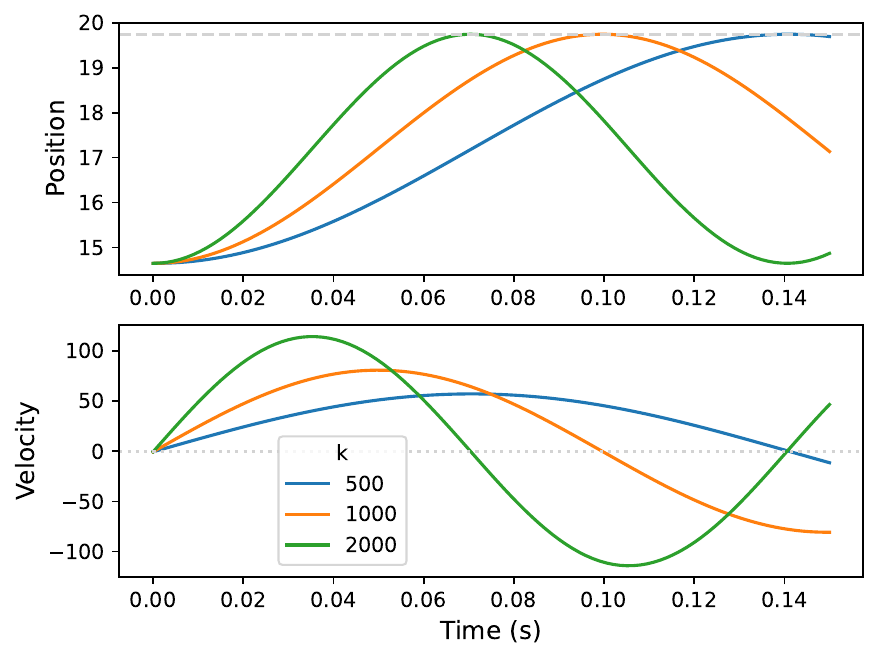}
\includegraphics[scale=0.5]{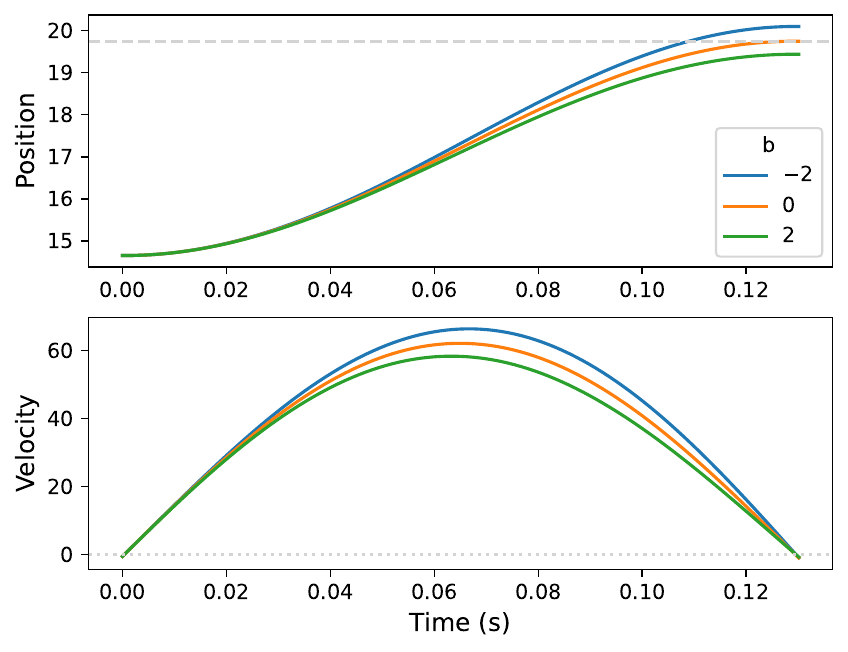}
\caption{The effect of varying $k$ on time-to-target achievement when $x_{0} = 14.65, T = 19.75, b=0$. The dashed line in the upper panel represents the value of $T$. (left). The effect of varying $b$ for the same trajectory, where $k = 592.74$ (right).}
\label{plot:secondorder_bk}
\end{figure}

Given the above, how are the appropriate parameters determined for a given duration? Figure \ref{plot:secondorder_sindy} shows simulated trajectories based on the best-fitting model, where $x_{0} = 14.65, T = 19.75$. The left panel of Figure \ref{plot:secondorder_bk} shows how changing $k \in \{500, 1000, 2000\}$ results in the target being met at shorter durations as $k$ is increased (when $b$ = 0). The system oscillates after target achievement, due to the system being undamped, so the gesture must be deactivated at target achievement. The right panel of Figure \ref{plot:secondorder_bk} shows that, for this example, target achievement occurs when $b = 0$; when $b$ is positive the trajectory is damped and undershoot occurs, when $b$ is negative overshoot occurs. We note that the version of the model with $\frac{k}{2}(T + x_{0})$ represents a specific case where the intended target is met, velocity trajectories are symmetrical, and the dynamics are highly linear.

In summary, the best linear model appears to be an under-damped harmonic oscillator. The model can be slightly improved by adding $\frac{k}{2}(T + x_{0})$, which relates the target to the initial condition. This adds some degree of complexity, largely in the dynamics around the equilibrium position, but note that this is not substantial and it fundamentally only involves adding a constant term to the standard harmonic oscillator model. It is worth further probing the theoretical implications of such a model, but note that recent advances in nonlinear modelling of speech gestures also incorporate information on initial position and movement distance directly into the differential equation \citep{kirkham2025}.

\subsection{Adding complexity: second-order nonlinear model}
\label{sec:models_secondorder_cubic}

\begin{figure}[h]
\centering
\includegraphics[scale=0.6]{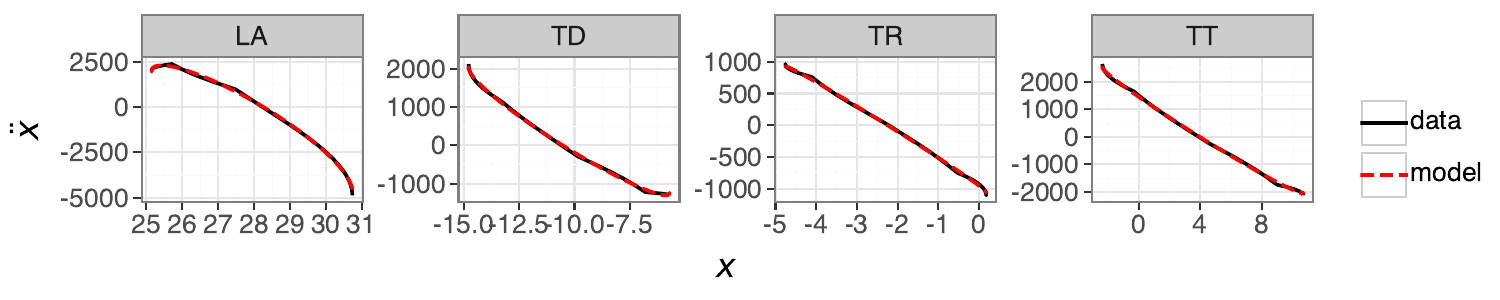}
\includegraphics[scale=0.6]{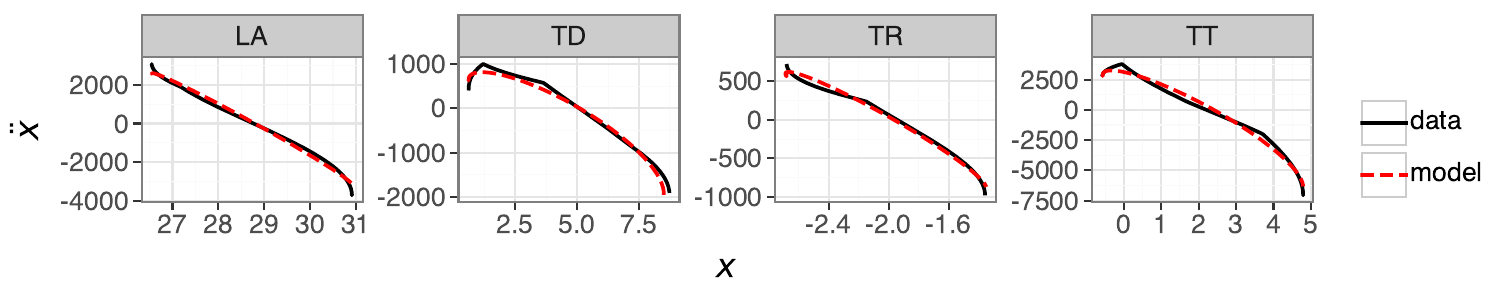}
\includegraphics[scale=0.6]{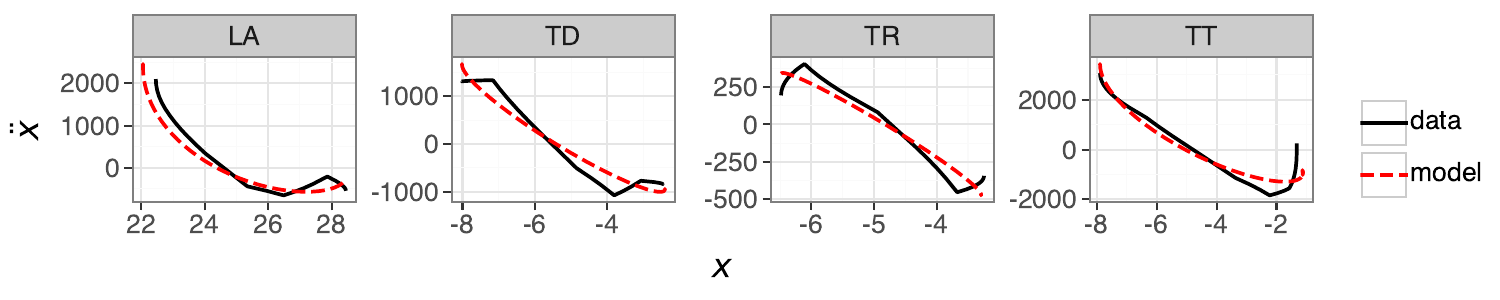}
\includegraphics[scale=0.6]{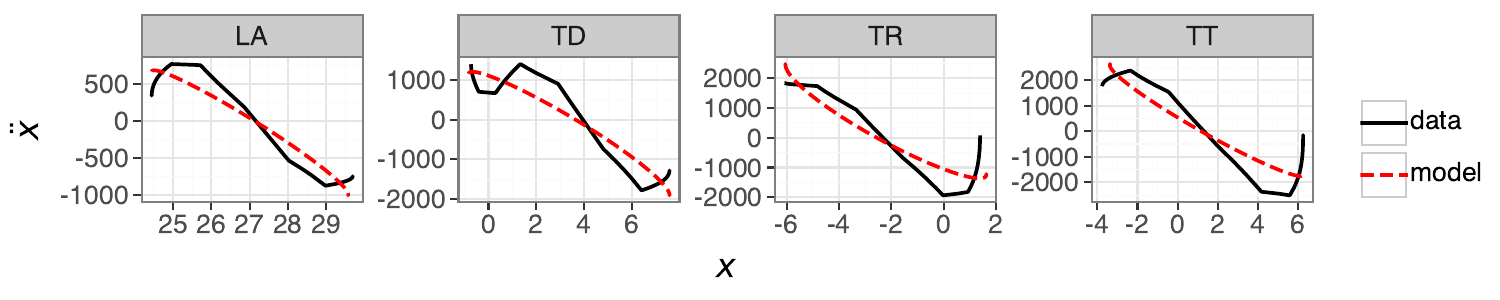}
\caption{Hooke portraits for second-order linear models with the highest score (top row), median score (second row), 5th percentile score (third row) and 1st percentile score (fourth row) for each articulatory variable.}
\label{plot:secondorder_hooke}
\end{figure}

Section \ref{sec:models_secondorder} shows that a linear model can capture movement dynamics with high accuracy, but recall that the selected examples were the best-fitting models. It should not be a surprise that a linear model accurately models dynamics that show strong linear signatures. How representative is this of the second-order model's overall performance? Figure \ref{plot:secondorder_hooke} repeats the Hooke portrait for the best-scoring models (top row) and compares this with median-scoring models (second row), 5th percentile scoring models (third row), and 1st percentile scoring models (fourth row) for each articulatory variable in the linear model. The 1st and 5th percentile models are those where only 1\% or 5\% of tokens are lower-scoring than these models. Note that the median scoring models the $R^2$ values range between $[0.989, 0.992]$, the 5th percentile models are in the range $[0.936, 0.965]$, and the 1st percentile models are in the range $[0.849, 0.915]$ . Recall that these scores are based on position and velocity, not acceleration, which is why they look more optimistic about the model's performance compared with the Hooke portraits. The median model is a good approximation of the data, while the 5th and 1st decile models struggle with the greater nonlinearity between position and acceleration. Note that this failure of the model is much clearer in the relationship between higher derivatives, because the comparison of position and velocity trajectories for these tokens do not obviously reveal this behaviour.

The lower scoring models in Figure \ref{plot:secondorder_hooke} correspond to cases where there is greater nonlinearity in the relation between position and acceleration. This indicates anharmonicity and is not within the scope of a linear harmonic oscillator \citep{sorensen-gafos2016}. In summary, while the first-order model can capture nonlinearity in the Hooke portraits, it over predicts the extent of nonlinearity and fails to adequately model the empirical characteristics of most of the data. In contrast, a second-order linear model is a better empirical fit, but lacks the ability to capture the nonlinearity present in some of the data. It stands to reason that a second-order nonlinear model should combine the strengths of the two approaches.

\begin{figure}[h]
\centering
\includegraphics[scale=0.6]{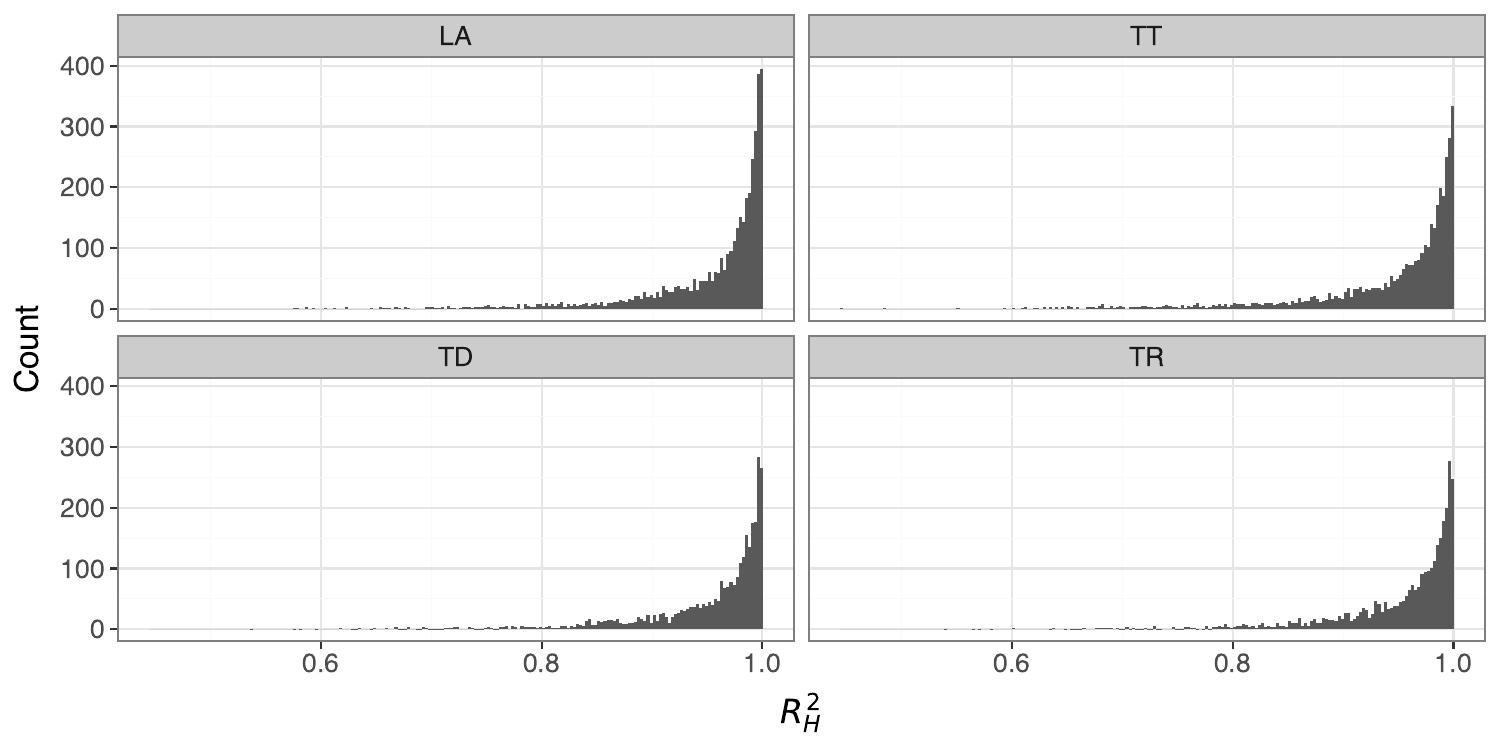}
\caption{Distribution of $R_{H}^2$ values for each sensor, based on a by-token linear regression of position against acceleration in the empirical trajectory data.}
\label{plot:hooke_linearity}
\end{figure}

Thus far, the results suggest a role for a nonlinear term in the second-order model. How extensive is nonlinearity in the data? Figure \ref{plot:hooke_linearity} shows the degree of linearity in the Hooke portrait for all trajectories, based on $R^2$ values from by-trajectory linear regression fits between position and acceleration \citep{mottet-bootsma1999}. Note that these $R^2$ values only correspond to the data and capture the linear fit between position and acceleration; they do not refer to any fit between the second-order model predictions and the data. To disambiguate, we subsequently refer to $R^2$ values for the Hooke portraits as $R_{H}^2$.

It is obviously challenging to pose a specific $R_{H}^2$ value that indicates nonlinearity in the Hooke portrait, as opposed to an under-performing fit due to measurement or segmentation errors. The lower row of Figure \ref{plot:hooke_linearity} (1st percentile) provides sufficient context, however, because the empirical trajectories here are clearly nonlinear, with associated $R_{H}^2$ values of 0.9 (LA), 0.85 (TD), 0.82 (TR) and 0.83 (TT). This suggests that $R_{H}^2 \leq 0.9$ certainly indicates substantial nonlinearity that is outside of the scope of a linear model. Across all trajectories, an average of 69\% of tokens across articulatory variables have $R_{H}^2 > 0.95$, whereas 15\% of tokens have $R^2 < 0.9$. From this, we can assume that approximately 30\% of tokens have a substantial degree of nonlinearity and that around half of these show extensive nonlinearity. Note that correlations between $R_{H}^2$ and spatial displacement are $r < 0.1$ for all articulatory variables. This suggests that nonlinearity does not straightforwardly interact with the magnitude of spatial displacement, as is predicted by a nonlinear model \citep{sorensen-gafos2016}, although see \citet{kirkham2025} for a nonlinear model where these dynamics are optionally under the control of the speaker.

\begin{figure}[h]
\centering
\includegraphics[scale=0.6]{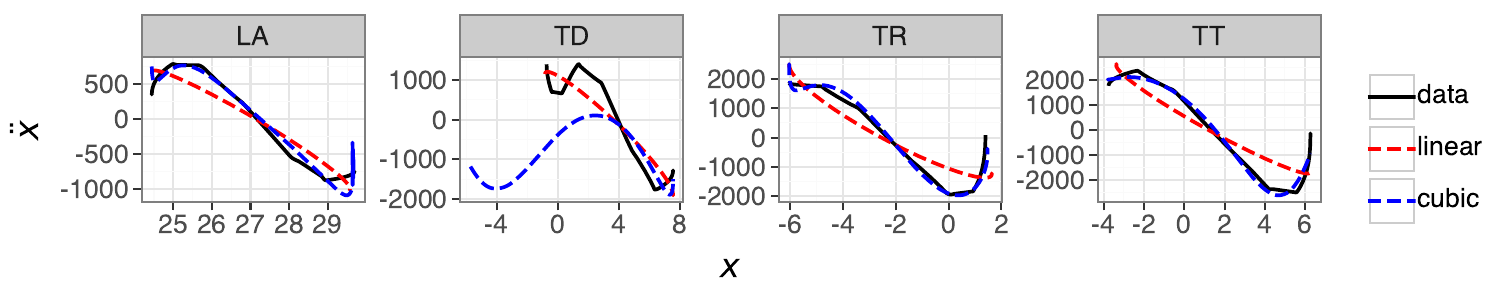}
\caption{Hooke portraits for linear and cubic second-order models. Note that the models for LA and TT contain $x^3$ and $\dot{x}^3$, whereas TD and TR only contain $x^3$.}
\label{plot:secondorder_hooke_cubic}
\end{figure}

That the majority of tokens show linearity in the Hooke portrait is why the linear model in Section \ref{sec:models_secondorder} shows good performance, but this analysis makes clear that a nonlinear term is required in order to account for the full dynamics of articulatory trajectories. In this case, the Hooke portraits reveal that some of the additional complexity in the higher polynomial libraries from Section \ref{subsec:second_order} may be warranted. Figure \ref{plot:secondorder_hooke_cubic} shows the same Hooke portraits for the 1st percentile trajectory fits as in Figure \ref{plot:secondorder_hooke} (lower row), but with with an additional cubic term in the feature library. Models were fitted using the same SINDy SR3 algorithm, but with a custom library comprising linear terms and a cubic term. The cubic term was initially constrained to position-only, but the LA and TT models required both $x^3$ and $\dot{x}^3$ in order to improve on the linear model. This is not necessarily unusual for models of human movement \citep{beek-beek1988, schoener1990} and there may be an advantage to the inclusion of nonlinear velocity terms more generally, especially for modelling qualitatively distinct movement dynamics, such as limit cycles \citep{kuberski-gafos2023}.

The nonlinear model clearly provides a better fit than the linear model for trajectories in Figure \ref{plot:secondorder_hooke_cubic}, with the exception of TD, where SINDy fails to find an optimal model. This suggests that the anharmonicity indicated by nonlinear relations between position and acceleration is well within the scope of a nonlinear model, as shown by \citet{sorensen-gafos2016}. It is clear that a single cubic term on position $x^3$ can provide an excellent nonlinear fit for the TR variable in Figure \ref{plot:secondorder_hooke_cubic}, while the poor fit for TD and presence of $\dot{x}^3$ for LA and TT are more likely to represent either (i) fitting to errors in the data; or (ii) a failure of the SR3 algorithm to find an optimal solution.

Parameterization of the cubic model is clearly more challenging than the linear model and further tests showed that the SR3 algorithm failed to find an optimal fit for a cubic model in many cases, despite the empirical data clearly being within the model's scope. Indeed, a cubic model cannot be easily derived by taking the SINDy coefficients for the linear model and adding a cubic term. For example, the linear coefficient values in the discovered linear and cubic models differ substantially, such that the linear model's damping coefficient indicates very weak damping, whereas the cubic model involves more substantial (but still sub-critical) damping. One potential source of difficulty in model fitting may be the complexity of the cubic coefficient, which can take on a very wide range of values depending on movement amplitude, even if movement characteristics are otherwise similar. \citet{kirkham2025} outlines a method for scaling the cubic coefficient by (actual or potential) movement amplitude. This constrains the parameter search to the range [0, 1), which should improve model fitting and interpretability. It may also be the case, as discussed above, that a cubic term on velocity is necessarily to fully encompass the range of speech movements, which remains an open line of inquiry for future research.

\subsection{Interim summary}

In summary, we have explored and interpreted three candidate models. The first-order model is nonlinear, but analysis of phase and Hooke portraits revealed the insufficiency of this model for capturing the full dynamics of the trajectories. We show that second-order linear model is a good approximation of the majority of the data, accurately capturing the dynamics for around 2/3 of trajectories, but there is significant nonlinearity in the Hooke portrait for around one third of trajectories. This is to be expected given our liberal velocity threshold, as \citet{kuberski-gafos2023} show that thresholded segmentation can lead to under-estimates of the extent of nonlinearity. It should be noted that we still find considerable evidence of quasi-linear relations in the Hooke portrait, but a straightforward comparison between studies is challenging. For example, \citet{kuberski-gafos2023} only examine closing movements (whereas we examine opening and closing movements) and they only examine repeated syllables at different metronome rates (whereas we include a wider range of speech materials). Regardless, the presence of nonlinearity points towards the need for greater model complexity, which can be adequately captured with the addition of a cubic term. 

Recall that in all cases the models are only minimally damped, which deviates from previous models that assume critical damping. Specifically, a perfectly symmetrical velocity trajectory can be achieved when the linear model is undamped, where the value of $k$ determines time-to-target achievement, but this requires a mechanism to deactivate the gesture upon reaching the target. Finally, a nonlinear model is clearly required to account for around one third of the data. This suggests that the movement dynamics of speech are fundamentally nonlinear, even though the nonlinear force may be minimal in some cases. In summary, the nonlinear model that best captures the full range of variation in the current data corresponds to a version of \citet{sorensen-gafos2016} model without critical damping

\begin{equation}
\label{eq:sg16_repeat}
\ddot{x} + a(t)[b\dot{x} + k(x-T) - d(x-T)^3] = 0.
\end{equation}

A more complex version of this model optionally transform the linear damping term $b\dot{x}$ into a nonlinear damping force $b\dot{x}^3$

\begin{equation}
\label{eq:sg16_v3}
\ddot{x} + a(t)[b\dot{x}^3 + k(x-T) - d(x-T)^3] = 0.
\end{equation}

In both cases, gestural dynamics are autonomous during activation, where $a(t)$ is step activation

\begin{equation}
a(t) = 
\begin{cases} 
    1, & t \in [t_{a}, t_{b}],\\ 
    0, & \text{otherwise.}
\end{cases}
\label{eq:rect_repeat}
\end{equation}

The following section now discusses the theoretical implications of these models, as well as limitations and prospects for model discovery in cognitive science.

\section{Discussion}
\label{sec:discussion}

\subsection{An evaluation of discovered models}

The first-order model initially shows good accuracy, in the sense that it provides an approximate fit to position and velocity trajectories. At first glance, any issues with the first-order model may appear to concern small deviations in fitting accuracy, but an analysis of the Hooke portraits reveals more fundamental issues. Specifically, the first-order model fails to accurately reproduce the relations between position and acceleration. In other words, what might look like minor differences in quantitative accuracy for position and velocity estimation is actually a failure to capture fundamental dynamics of the system. This renders a first-order autonomous model insufficient for capturing articulatory dynamics and highlights the importance of exploring the scope of the discovered model's predictions, rather than only fitting to empirical trajectories.

The second-order model shows better empirical fits than the first-order model and the highest-performing cases show excellent fits in the phase and Hooke portraits. Under the linear interpretation we find that the target, stiffness and damping terms need to be tuned in just the right way to meet the target in the specified time interval. The biggest differences between the second-order model presented here and previous task dynamic models are (1) we relax the critical damping constraint; (2) target achievement is non-asymptotic. In other words, the specific target value is achieved, rather than the system becoming asymptotically close.

A second-order linear model with minimal damping and a reformulation of the target term is capable of accurately modelling the majority of articulatory trajectories. As with the first-order model, however, cases of under-performance do not simply reflect minor issues in quantitative fit or data quality issues. Instead, the Hooke portraits show that the linear model is incapable of capturing the observable nonlinearity between position and acceleration, which occurs in approximately one third of trajectories. This motivated a second-order nonlinear model with a cubic term \citep{sorensen-gafos2016}, which is typically under-damped rather than critically damped. While the nonlinear model is significantly more difficult to parameterize, it is able to accurately model the observed nonlinearities. This confirms that the articulatory dynamics of speech are fundamentally nonlinear and adds considerable support to the \citet{sorensen-gafos2016} model of articulatory control.

\subsection{Autonomous dynamics and beyond}
\label{subsec:discussion_autonomous}

\citet{beek-beek1988} outline a typology of dynamical models of rhythmic movement, including (1) simple linear models with complex external forcing; (2) nonlinear autonomous models; (3) nonlinear models with minimal external forcing. Approach (1) is a relatively common theme in computational and modular models of mind, where movement models are simple and most of the work is offloaded to a complex forcing function. This essentially reflects a view in which physical movement is the implementation stage of highly complex cognitive processing. In contrast, the present analysis has focused mainly on the second of Beek \& Beek's \citeyearpar{beek-beek1988} typology, if we also allow the inclusion of some good-performing linear models. Indeed, we confirm that the intrinsic dynamics of articulatory trajectories can be modelled without any explicit time-dependence across a large database of 13,742 segmented trajectories. Our approach has been to assume very simple step-activation driving of the system, with instantaneous change in parameters at specific landmarks. However, this rather simple driving mechanism is likely insufficient, given that it does not explain how the gestural system is driven from one target to another.

A more comprehensive dynamical model would likely occupy the third category of Beek \& Beek's \citeyearpar{beek-beek1988} typology, with nonlinear gestural dynamics and an external forcing function $F(t)$ that drives between system states. A clear example concerns the forces that drive the initiation and termination of gestures. This is particularly pertinent to the present study, as the linear second-order model is close to undamped and the coefficients discovered for a cubic second-order model are almost never critically-damped (but significantly more damped than the linear model). This means that both models require some form of gestural suppression mechanism, otherwise the system will inherently oscillate around the target. One proposal is state feedback on target achievement \citep{tilsen2022, burroni-tilsen2022}, where speakers use a combination of internal and external feedback to open and close a gestural gating function. \citet{parrell-etal2019} also outline a model of hierarchical state feedback control, which combines task dynamics with nonlinear state estimation.

An alternative proposal rejects the idea that gestures have bounded activation intervals altogether, instead casting gestures as always active but varying in their force on the vocal tract \citep{tilsen2018}. Under this view, movement preparation is represented as a dynamic neural planning field \citep{erlhagen-schoener2002, schoener2020}, where articulatory control variables correspond to a parameter field. Gestures act as inputs to the planning fields, with the field's activation centroid determining the parameter value. In this view, the parameters of minimal dynamical models are re-cast as dynamic neural fields that generate continuous values from activation across neural populations \citep{kirov-gafos2007, roon-gafos2016, tilsen2019, stern-shaw2023, kirkham-strycharczuk2024}. The task of model discovery thus becomes uncovering the dynamics of neural fields that continuously parameterize gestural systems. This view clearly situates gestural systems as explicitly non-autonomous, because parameters are no longer constant and are constantly being fed by movement planning fields.

 At this juncture, it should be noted that the distinction between autonomous and non-autonomous does not necessarily have to correspond to a distinction between intrinsic versus extrinsic timing \citep{fowler1980}, where the latter is associated with a central clock or time-keeper \citep[e.g.][]{turk-shattuck-hufnagel2020}. It is self-evident that no living system is autonomous, because interaction is a signature of life \citep{supruneko-etal2013}, but the present study shows that the intrinsic dynamics of independent gestures can be adequately modelled as autonomous systems. That said, even if non-autonomous models turn out to be the correct direction, any time-varying parameters may still be a consequence of coupling to other dynamical mechanisms with their own intrinsic dynamics, rather than a central time-keeper. This is consistent with accounts of interoceptive rhythms in brain and body that can be coupled with the surrounding environment \citep[e.g.][]{engelen-etal2023}. To this end, the most productive perspective may be a view of brain and behaviour as a complex multi-scale system with interacting, coupled, and emergent dynamics \citep[e.g.][]{tilsen2009, favela2024, goheen-etal2024, kluger-etal2024, senkowski-engel2024}. It is the task of future research understand the nature of these dynamics of their coupling relations.

\subsection{Prospects for data-driven model discovery in cognitive science}

A major aim of the present work was to discover dynamical models from data. In doing so, we discovered some new models, but also leveraged extant models in improving on these discoveries, such as the second-order nonlinear model in \citet{sorensen-gafos2016}. It is worth noting that articulatory trajectories are comparably easier to model than some other dynamical mechanisms in the cognitive sciences, but there remain a range of areas in which data-driven model discovery represents a promising direction.

For example, any model that is concerned with the relationship between discrete categories and their physical realisation must take seriously both signed and spoken languages. While there is a very small amount of work exploring the possibility of a task dynamics of signed languages \citep[e.g.][]{mertz-etal2024}, this represents a space where model discovery would be particularly useful as a starting point, given appropriate kinematic corpora on signing. Debates also abound on the appropriate dynamical representation of various kinds of disordered speech, such as whether the addition of noise is sufficient \citep{mucke-etal2024}, or whether a different set of compensatory dynamics are involved.

While physical movements are obvious candidates for model discovery, this does not preclude data-driven discovery of higher-level cognitive processes. This includes decision making, perception and working memory, all of which can be cast as relatively simple dynamical models \citep{schoener-etal2016} and are, therefore, potential areas where data-driven model discovery could be productive. More broadly, it is likely that moving beyond intrinsic speech dynamics towards a larger agent-environment system will also require a broader reconceptualization of the dynamical models we deploy, with an open system tending towards massive degrees of interactivity and complexity. This is likely to require a focus on how interacting task demands constrain the dynamics of brain and behaviour, rather than attempting to model specific cognitive processes \citep{iskarous2016, nau-etal2024}.

\subsection{Limitations and future research}
 
 A productive avenue for future research would be more extensive comparisons between models \citep[e.g.][]{elie-etal2023}, especially on languages other than English \citep{geissler-nellakra2024}. This is important given the emergence of new nonlinear models \citep{stern-shaw2024} and ongoing developments in theories of articulatory control \citep{tilsen2016, tilsen2018, tilsen2020}. It will be particularly important to compare the qualitative predictions made by different models in order to distinguish quantitative fit from qualitative adequacy (e.g. via phase and Hooke portraits). In future research, we also hope to examine bi-directional coupling between the dynamics of neural fields and the dynamics of nonlinear gestural models, which are likely mediated by dynamical feedback mechanisms \citep[e.g.][]{parrell-etal2019, tilsen2022}. There is extensive scope for model development in these areas, but attention should also be directed towards developing rigorous ways of testing the predictions of different models.
 
A limitation of the current approach is temporal segmentation of gestures, and the method's reliance on accurate segmentation. We segmented signals at velocity zero-crossings, based on well-understood characteristics of skilled movement dynamics. If two gestures overlap then we are not able to distinguish them, or estimate the parameters of each gesture separately. But this conceptualization is likely to be a gross simplification of the actual dynamics of gestures. As discussed above, \citet{tilsen2019} outlines a model in which gestures are always active, but most are sub-threshold at any point in time. In terms of model discovery, this becomes a significantly more complex task, but it is possible that a combination of sparse symbolic regression, predictive control algorithms, and neural networks may prove a fruitful avenue for model discovery in this area \citep{kaiser-etal2018, tilsen2020}.

\section{Conclusions}

Discovering the dynamics that govern brain and behaviour is a major challenge in the cognitive sciences. We have demonstrated one approach to meeting this challenge, applied to articulatory movement dynamics in spoken language. Building upon decades of research in task dynamics and Articulatory Phonology, combined with recent developments in machine learning and equation discovery, we discover interpretable models directly from data. While a linear second-order equation accurately models around two thirds of trajectories in the data, a nonlinear (cubic) model is fundamentally necessary for accurately capturing the qualitative dynamics of speech movements. This supports the proposal that articulatory dynamics are well-modelled as a nonlinear autonomous system during periods of constant gestural activation. This leads us to propose that the discovered models represent the dynamical laws of motion that structure articulatory control in spoken language.

\section*{Acknowledgements}

Many thanks to the editor Rick Dale and three reviewers for their constructive feedback, as well as audiences at BAAP 2024 (Cardiff), ISSP 2024 (Autrans), and Time Series Analysis of Noisy Data 2024 (Lancaster). I nonetheless accept full responsibility for any remaining shortcomings. This research was funded by UKRI/AHRC fellowship AH/Y002822/1. Figure \ref{fig:vt_spring_model} was created in BioRender: \url{https://BioRender.com/c94z204}.


\begin{thebibliography}{93}
\expandafter\ifx\csname natexlab\endcsname\relax\def\natexlab#1{#1}\fi
\expandafter\ifx\csname url\endcsname\relax
  \def\url#1{{\tt #1}}\fi
\expandafter\ifx\csname urlprefix\endcsname\relax\def\urlprefix{URL }\fi

\bibitem[{Abakarova, Fuchs \& Noiray(2022)D.~Abakarova, S.~Fuchs \&
  A.~Noiray}]{abakarova-etal2022}
Abakarova, Dzhuma, Susanne Fuchs \& Aude Noiray. 2022. Developmental changes in
  coarticulation degree relate to differences in articulatory patterns: An
  empirically grounded modeling approach. {\em Journal of Speech, Language, and
  Hearing Research\/} 65(9), 3276--3299.

\bibitem[{Abraham \& Shaw(1992)R.~H. Abraham \& C.~D. Shaw}]{abraham-shaw1992}
Abraham, Ralph~H. \& Christopher~D. Shaw. 1992. {\em Dynamics: The Geometry of
  Behavior\/}. Redwood City, CA: Addison-Wesley.

\bibitem[{Anderson(1972)P.~Anderson}]{anderson1972}
Anderson, P.W. 1972. More is different: Broken symmetry and the nature of the
  hierarchical struture of science. {\em Science\/} 177(4047), 393--396.

\bibitem[{Barrett(2011)L.~Barrett}]{barrett2011}
Barrett, Louise. 2011. {\em Beyond the Brain: How Body and Environment Shape
  Animal and Human Minds\/}. Princeton, NJ: Princeton University Press.

\bibitem[{Beek \& Beek(1988)P.~Beek \& W.~Beek}]{beek-beek1988}
Beek, P.J. \& W.J. Beek. 1988. Tools for constructing dynamical models of
  rhythmic movement. {\em Human Movement Science\/} 7(2--4), 301--342.

\bibitem[{Birkholz, Kroger \& Neuschaefer-Rube(2011)P.~Birkholz, J.~Kroger \&
  C.~Neuschaefer-Rube}]{birkholz-etal2011}
Birkholz, Peter, J.~Kroger, Bernd \& Christiane Neuschaefer-Rube. 2011.
  Model-based reproduction of articulatory trajectories for consonant-vowel
  sequences. {\em {IEEE} Transactions on Audio, Speech, and Language
  Processing\/} 19(5), 1422--1433.

\bibitem[{Browman \& Goldstein(1992)C.~P. Browman \&
  L.~Goldstein}]{browman-goldstein1992}
Browman, Catherine~P. \& Louis Goldstein. 1992. Articulatory phonology: An
  overview. {\em Phonetica\/} 49(3-4), 155--180.

\bibitem[{Browman \& Goldstein(1986)C.~P. Browman \& L.~M.
  Goldstein}]{browman-goldstein1986}
Browman, Catherine~P. \& Louis~M. Goldstein. 1986. Towards an articulatory
  phonology. {\em Phonology\/} 3(1), 219--252.

\bibitem[{Brunton \& Kutz(2022)S.~L. Brunton \& J.~N. Kutz}]{brunton-kutz2022}
Brunton, Steven~L. \& J.~Nathan Kutz. 2022. {\em Data-Driven Science and
  Engineering: Machine Learning, Dynamical Systems, and Control\/}. Cambridge:
  Cambridge University Press.

\bibitem[{Brunton, Proctor \& Kutz(2016)S.~L. Brunton, J.~L. Proctor \& J.~N.
  Kutz}]{brunton-etal2016}
Brunton, Steven~L., Joshua~L. Proctor \& J.~Nathan Kutz. 2016. Discovering
  governing equations from data by sparse identification of nonlinear dynamical
  systems. {\em Proceedings of the National Academy of Sciences\/} 113(15),
  3932--3937.

\bibitem[{Burroni \& Tilsen(2022)F.~Burroni \& S.~Tilsen}]{burroni-tilsen2022}
Burroni, Francesco \& Sam Tilsen. 2022. The online effect of clash is
  durational lengthening, not prominence shift: Evidence from {I}talian. {\em
  Journal of Phonetics\/} 91(101124).

\bibitem[{Burton(1965)T.~A. Burton}]{burton1965}
Burton, T.~A. 1965. The generalized {L}i\'{e}nard equation. {\em Journal of the
  Society for Industrial and Applied Mathematics Series A Control\/} 3(2),
  223--230.

\bibitem[{Byrd \& Saltzman(1998)D.~Byrd \& E.~Saltzman}]{byrd-saltzman1998}
Byrd, Dani \& Elliot Saltzman. 1998. Intragestural dynamics of multiple
  prosodic boundaries. {\em Journal of Phonetics\/} 26(2), 173--199.

\bibitem[{Byrd \& Saltzman(2003)D.~Byrd \& E.~Saltzman}]{byrd-saltzman2003}
Byrd, Dani \& Elliot Saltzman. 2003. The elastic phrase: modeling the dynamics
  of boundary-adjacent lengthening. {\em Journal of Phonetics\/} 31(2),
  149--180.

\bibitem[{Carello, Turvey, Kugler et~al.(1984)C.~Carello, M.~A. Turvey, P.~N.
  Kugler \& R.~E. Shaw}]{carello-etal1984}
Carello, Claudia, Michael~A. Turvey, Peter~N. Kugler \& Robert~E. Shaw. 1984.
  Inadequacies of the computer metaphor. In Michael~S. Gazzaniga (ed.) {\em
  Handbook of Cognitive Neuroscience\/}, 229--248, New York, NY: Plenum Press.

\bibitem[{Champion, Zheng, Aravkin et~al.(2020)K.~Champion, P.~Zheng, A.~Y.
  Aravkin, S.~L. Brunton \& J.~N. Kutz}]{champion-etal2020}
Champion, Kathleen, Peng Zheng, Aleksandr~Y. Aravkin, Steven~L. Brunton \&
  J.~Nathan Kutz. 2020. A unified sparse optimization framework to learn
  parsimonious physics-informed models from data. {\em {IEEE} Access\/} 8,
  169259--169271.

\bibitem[{Chartier, Anumanchipalli, Johnson et~al.(2018)J.~Chartier, G.~K.
  Anumanchipalli, K.~Johnson \& E.~Chang}]{chartier-etal2018}
Chartier, Josh, Gopala~K. Anumanchipalli, Keith Johnson \& Edward Chang. 2018.
  Encoding of articulatory kinematic trajectories in human speech sensorimotor
  cortex. {\em Neuron\/} 98(5), 1042--1054.

\bibitem[{Chemero(2009)A.~Chemero}]{chemero2009}
Chemero, Anthony. 2009. {\em Radical Embodied Cognitive Science\/}. Cambridge,
  MA: MIT Press.

\bibitem[{Chicco, Warrens \& Jurman(2021)D.~Chicco, M.~J. Warrens \&
  G.~Jurman}]{chicco-etal2021}
Chicco, Davide, Matthijs~J. Warrens \& Giuseppe Jurman. 2021. The coefficient
  of determination {R}-squared is more informative than {SMAPE}, {MAE}, {MAPE},
  {MSE} and {RMSE} in regression analysis evaluation. {\em PeerJ Computer
  Science\/} 7(e623), 1--24.

\bibitem[{Chomsky \& Halle(1968)N.~Chomsky \& M.~Halle}]{chomsky-halle1968}
Chomsky, Noam \& Morris Halle. 1968. {\em The Sound Pattern of {E}nglish\/}.
  New York, NY: Harper and Row.

\bibitem[{Dale \& Bhat(2018)R.~Dale \& H.~S. Bhat}]{dale-bhat2018}
Dale, Rick \& Harish~S. Bhat. 2018. Equations of mind: Data science for
  inferring nonlinear dynamics of socio-cognitive systems. {\em Cognitive
  Systems Research\/} 52, 275--290.

\bibitem[{{de Silva}, Champion, Quade et~al.(2020)B.~M. {de Silva},
  K.~Champion, M.~Quade, J.-C. Loiseau, J.~N. Kutz \& S.~L.
  Brunton}]{desilva-etal2020}
{de Silva}, Brian~M., Kathleen Champion, Markus Quade, Jean-Christophe Loiseau,
  J.~Nathan Kutz \& Steven~L. Brunton. 2020. {PySINDy}: A {P}ython package for
  the sparse identification of nonlinear dynamical systems from data. {\em
  Journal of Open Source Software\/} 5(49), 2104.

\bibitem[{Elie, Lee \& Turk(2023)B.~Elie, D.~N. Lee \& A.~Turk}]{elie-etal2023}
Elie, Benjamin, David~N. Lee \& Alice Turk. 2023. Modeling trajectories of
  human speech articulators using general {T}au theory. {\em Speech
  Communication\/} 151, 24--38.

\bibitem[{Engelen, Solc\`{a} \& Tallon-Baudry(2023)T.~Engelen, M.~Solc\`{a} \&
  C.~Tallon-Baudry}]{engelen-etal2023}
Engelen, Tahn\'{e}e, Macro Solc\`{a} \& Catherine Tallon-Baudry. 2023.
  Interoceptive rhythms in the brain. {\em Nature Neuroscience\/} 26,
  1670--1684.

\bibitem[{Erlhagen \& Sch\"{o}ner(2002)W.~Erlhagen \&
  G.~Sch\"{o}ner}]{erlhagen-schoener2002}
Erlhagen, Wolfram \& Gregor Sch\"{o}ner. 2002. Dynamic field theory of movement
  preparation. {\em Psychological Review\/} 109(3), 545--572.

\bibitem[{Favela(2024)L.~H. Favela}]{favela2024}
Favela, Luis~H. 2024. {\em The Ecological Brain: Unifying the Sciences of
  Brain, Body, and Environment\/}. New York, NY: Routledge.

\bibitem[{Fodor(1975)J.~A. Fodor}]{fodor1975}
Fodor, Jerry~A. 1975. {\em The Language of Thought\/}. Cambridge, MA: Harvard
  University Press.

\bibitem[{Fowler(1980)C.~A. Fowler}]{fowler1980}
Fowler, Carol~A. 1980. Coarticulation and theories of extrinsic timing. {\em
  Journal of Phonetics\/} 8(1), 113--133.

\bibitem[{Gafos(2006)A.~I. Gafos}]{gafos2006}
Gafos, Adamantios~I. 2006. Dynamics in grammar. In Louis Goldstein, D.H. Whalen
  \& Catherine~T. Best (eds.) {\em Laboratory Phonology 8: Varieties of
  Phonological Competence\/}, 51--79, Berlin: Mouton de Gruyter.

\bibitem[{Gafos \& Benus(2006)A.~I. Gafos \& S.~Benus}]{gafos-benus2006}
Gafos, Adamantios~I. \& Stefan Benus. 2006. Dynamics of phonological cognition.
  {\em Cognitive Science\/} 30(5), 905--943.

\bibitem[{Geissler \& Nellakra(2024)C.~Geissler \&
  J.~Nellakra}]{geissler-nellakra2024}
Geissler, Christopher \& Jyothiraditya Nellakra. 2024. Predicting articulatory
  landmarks with critically-damped oscillators and {G}eneral {T}au theory. {\em
  Proc. {ISSP} 2024 -- 13th International Seminar on Speech Production\/}
  173--176.

\bibitem[{Gibson(1979)J.~J. Gibson}]{gibson1979}
Gibson, James~J. 1979. {\em The Ecological Approach to Visual Perception\/}.
  Boston, MA: Houghton Mifflin.

\bibitem[{Goheen, Wolman, Angeletti et~al.(2024)J.~Goheen, A.~Wolman, L.~L.
  Angeletti, A.~Wolff, J.~A. Anderson \& G.~Northoff}]{goheen-etal2024}
Goheen, Josh, Angelika Wolman, Lorenzo~Lucherini Angeletti, Annemarie Wolff,
  John~A.E. Anderson \& Georg Northoff. 2024. Dynamic mechanisms that couple
  the brain and breathing to the external environment. {\em Communications
  Biology\/} 7(938), 1--11.

\bibitem[{Guenther(2016)F.~H. Guenther}]{guenther2016}
Guenther, Frank~H. 2016. {\em Neural Control of Speech\/}. Cambridge, MA: The
  MIT Press.

\bibitem[{Haken(1977)H.~Haken}]{haken1977}
Haken, Hermann. 1977. {\em Synergetics: An Introduction\/}. Berlin:
  Springer-Verlag.

\bibitem[{Haken, Kelso \& Bunz(1985)H.~Haken, J.~S. Kelso \&
  H.~Bunz}]{haken-etal1985}
Haken, Hermann, J.A.~Scott Kelso \& H.~Bunz. 1985. A theoretical model of phase
  transitions in human hand movements. {\em Biological Cybernetics\/} 51(5),
  347--356.

\bibitem[{Hooke(1678)R.~Hooke}]{hooke1678}
Hooke, Robert. 1678. {\em De Potentia Restitutiva, or of Spring: Explaining the
  Power of Springing Bodies\/}. London: John Martyn.

\bibitem[{Hume(1739)D.~Hume}]{hume1739}
Hume, David. 1739. {\em A Treatise of Human Nature\/}. London: John Noon.

\bibitem[{Iskarous(2016)K.~Iskarous}]{iskarous2016}
Iskarous, Khalil. 2016. Compatible dynamical models of environmental, sensory,
  and perceptual systems. {\em Ecological Psychology\/} 28(4), 295--311.

\bibitem[{Iskarous(2017)K.~Iskarous}]{iskarous2017}
Iskarous, Khalil. 2017. The relation between the continuous and the discrete: A
  note on the first principles of speech dynamics. {\em Journal of Phonetics\/}
  64, 8--20.

\bibitem[{Iskarous, Cole \& Steffman(2024)K.~Iskarous, J.~Cole \&
  J.~Steffman}]{iskarous-etal2024}
Iskarous, Khalil, Jennifer Cole \& Jeremy Steffman. 2024. A minimal dynamical
  model of intonation: Tone contrast, alignment, and scaling of {A}merican
  {E}nglish pitch accents as emergent properties. {\em Journal of Phonetics\/}
  101309(1--27).

\bibitem[{Kaiser, Kutz \& Brunton(2018)E.~Kaiser, J.~N. Kutz \& S.~L.
  Brunton}]{kaiser-etal2018}
Kaiser, Eurika, J.~Nathan Kutz \& Steven~L. Brunton. 2018. Sparse
  identification of nonlinear dynamics for model predictive control in the
  low-data limit. {\em Proceedings of the Royal Society A: Mathematical,
  Physical and Engineering Sciences\/} 474(2219), 1--25.

\bibitem[{Keating(1990)P.~Keating}]{keating1990}
Keating, Pat. 1990. The window model of coarticulation: articulatory evidence.
  In John Kingston \& Mary~E. Beckman (eds.) {\em Papers in Laboratory
  Phonology I\/}, 3--29, Cambridge: Cambridge University Press.

\bibitem[{Kelso(1995)J.~S. Kelso}]{kelso1995}
Kelso, J.A.~Scott. 1995. {\em Dynamic Patterns: The self-organization of brain
  and behavior\/}. Cambridge, MA: MIT Press.

\bibitem[{Kelso, Saltzman \& Tuller(1986)J.~S. Kelso, E.~L. Saltzman \&
  B.~Tuller}]{kelso-etal1986}
Kelso, J.A.~Scott, Elliot~L. Saltzman \& Betty Tuller. 1986. The dynamical
  perspective on speech production: data and theory. {\em Journal of
  Phonetics\/} 14(1), 29--59.

\bibitem[{Kirkham(2024)S.~Kirkham}]{kirkham2024}
Kirkham, Sam. 2024. Discovering dynamical models of speech using
  physics-informed machine learning. {\em Proc. ISSP 2024 -- 13th International
  Seminar on Speech Production\/} 185--188.

\bibitem[{Kirkham(2025)S.~Kirkham}]{kirkham2025}
Kirkham, Sam. 2025. Scaling laws for nonlinear dynamical models of articulatory
  control. {\em {JASA} Express Letters\/} 5(2), 1--7.

\bibitem[{Kirkham \& Strycharczuk(2024)S.~Kirkham \&
  P.~Strycharczuk}]{kirkham-strycharczuk2024}
Kirkham, Sam \& Patrycja Strycharczuk. 2024. A dynamic neural field model of
  vowel diphthongisation. {\em Proc. ISSP 2024 -- 13th International Seminar on
  Speech Production\/} 193--196.

\bibitem[{Kirov \& Gafos(2007)C.~Kirov \& A.~I. Gafos}]{kirov-gafos2007}
Kirov, Christo \& Adamantios~I. Gafos. 2007. Dynamic phonetic detail in lexical
  representations. {\em Proceedings of the 16th International Congress of
  Phonetic Sciences\/} 637--640.

\bibitem[{Kluger, Allen \& Gross(2024)D.~A. Kluger, M.~G. Allen \&
  J.~Gross}]{kluger-etal2024}
Kluger, Daniel~A., Micah~G. Allen \& Joachim Gross. 2024. Brain-body states
  embody complex temporal dynamics. {\em Trends in Cognitive Sciences\/} 28(8),
  695--698.

\bibitem[{Kr\"{o}ger, Schr\"{o}der \& Opgen-Rhein(1995)B.~J. Kr\"{o}ger,
  G.~Schr\"{o}der \& C.~Opgen-Rhein}]{kroger-etal1995}
Kr\"{o}ger, Bernd~J., Georg Schr\"{o}der \& Claudia Opgen-Rhein. 1995. A
  gesture-based dynamic model describing articulatory movement data. {\em
  Journal of the Acoustical Society of America\/} 98(4), 1878--1889.

\bibitem[{Kuberski \& Gafos(2023)S.~R. Kuberski \& A.~I.
  Gafos}]{kuberski-gafos2023}
Kuberski, Stephan~R. \& Adamantios~I. Gafos. 2023. How thresholding in
  segmentation affects the regression performance of the linear model. {\em
  {JASA} Express Letters\/} 3(095202), 1--8.

\bibitem[{{LaFolette}, Yuval, Schurr et~al.(2024)K.~{LaFolette}, J.~Yuval,
  R.~Schurr, D.~Melnikoff \& A.~Goldenberg}]{lafollette-etal2024}
{LaFolette}, Kyle, Janni Yuval, Roey Schurr, David Melnikoff \& Amit
  Goldenberg. 2024. Data driven equation discovery reveals non-linear
  reinforcement learning in humans. {\em Pre-print available at:
  https://doi.org/10.31234/osf.io/65jqh\/} .

\bibitem[{Mertz, Pagel, Turco et~al.(2024)J.~Mertz, L.~Pagel, G.~Turco \&
  D.~M\"{u}cke}]{mertz-etal2024}
Mertz, Justine, Lena Pagel, Giuseppina Turco \& Doris M\"{u}cke. 2024.
  Gradiency and categoriality in the prosodic modulation of {F}rench {S}ign
  {L}anguage: A kinematic approach using electromagnetic articulography. {\em
  Proceedings of Speech Prosody 2024\/} 851--855.

\bibitem[{Mottet \& Bootsma(1999)D.~Mottet \& R.~J.
  Bootsma}]{mottet-bootsma1999}
Mottet, Denis \& Reinoud~J. Bootsma. 1999. The dynamics of goal-directed
  rhythmical aiming. {\em Biological Cybernetics\/} 80(4), 235--245.

\bibitem[{M\"{u}cke, Roessig, Thies et~al.(2024)D.~M\"{u}cke, S.~Roessig,
  T.~Thies, A.~Hermes \& A.~Mefferd}]{mucke-etal2024}
M\"{u}cke, Doris, Simon Roessig, Tabea Thies, Anne Hermes \& Antje Mefferd.
  2024. Challenges with the kinematic analysis of neurotypical and impaired
  speech: Measures and models. {\em Journal of Phonetics\/} 102(101292), 1--14.

\bibitem[{Nalepka, Lamb, Kallen et~al.(2019)P.~Nalepka, M.~Lamb, R.~W. Kallen,
  K.~Shockley, A.~Chemero, E.~Saltzman \& M.~J. Richardson}]{nalepka-etal2019}
Nalepka, Patrick, Maurice Lamb, Rachel~W. Kallen, Kevin Shockley, Anthony
  Chemero, Elliot Saltzman \& Michael~J. Richardson. 2019. Human social motor
  solutions for human-machine interaction in dynamical task contexts. {\em
  Proceedings of the National Academy of Sciences\/} 116(4), 1437--1446.

\bibitem[{Nau, Schmid, Kaplan et~al.(2024)M.~Nau, A.~C. Schmid, S.~M. Kaplan,
  C.~I. Baker \& D.~J. Kravitz}]{nau-etal2024}
Nau, Matthias, Alexandra~C. Schmid, Simon~M. Kaplan, Chris~I. Baker \&
  Dwight~J. Kravitz. 2024. Centering cognitive neuroscience on task demands and
  generalization. {\em Nature Neuroscience\/} 1--12.

\bibitem[{Newton(1687)I.~Newton}]{newton1687}
Newton, Isaac. 1687. {\em Philosophiae naturalis principia mathematica\/}.
  London: Jussu Societatis Regi\ae\ ac Typis Joseph Streate.

\bibitem[{Noether(1918)E.~Noether}]{noether1918}
Noether, Emmy. 1918. Invariante variationsprobleme. {\em Nachrichten der
  K\"{o}niglichen Gesellschaft der Wissenschaften zu G\"{o}ttingen,
  Mathematisch-Physikalische Klasse\/} 235--257.

\bibitem[{Ostry \& Munhall(1985)D.~J. Ostry \& K.~G.
  Munhall}]{ostry-munhall1985}
Ostry, David~J. \& Kevin~G. Munhall. 1985. Control of rate and duration of
  speech movements. {\em Journal of the Acoustical Society of America\/} 77(2),
  640--648.

\bibitem[{Parrrell, Ramanarayanan, Nagarajan et~al.(2019)B.~Parrrell,
  V.~Ramanarayanan, S.~Nagarajan \& J.~Houde}]{parrell-etal2019}
Parrrell, Benjamin, Vikram Ramanarayanan, Srikantan Nagarajan \& John Houde.
  2019. The {FACTS} model of speech motor control: Fusing state estimation and
  task-based control. {\em {PLoS} Computational Biology\/} 15(9), 1--26.

\bibitem[{Port \& {van Gelder}(1998)R.~F. Port \& T.~{van
  Gelder}}]{port-vangelder1998}
Port, Robert~F. \& Timothy {van Gelder} (eds.) . 1998. {\em Mind as Motion:
  Explorations in the Dynamics of Cognition\/}. Cambridge, MA: MIT Press.

\bibitem[{Roon \& Gafos(2016)K.~D. Roon \& A.~I. Gafos}]{roon-gafos2016}
Roon, Kevin~D. \& Adamantios~I. Gafos. 2016. Perceiving while producing:
  Modeling the dynamics of phonological planning. {\em Journal of Memory and
  Language\/} 89(2), 222--243.

\bibitem[{Saltzman \& Kelso(1987)E.~Saltzman \& J.~S.
  Kelso}]{saltzman-kelso1987}
Saltzman, Elliot \& J.A.~Scott Kelso. 1987. Skilled actions: a task-dynamic
  approach. {\em Psychological Review\/} 94(1), 84--106.

\bibitem[{Saltzman \& Munhall(1989)E.~Saltzman \& K.~G.
  Munhall}]{saltzman-munhall1989}
Saltzman, Elliot \& Kevin~G. Munhall. 1989. A dynamical approach to gestural
  patterning in speech production. {\em Ecological Psychology\/} 1(4),
  333--382.

\bibitem[{Schmidt \& Lipson(2009)M.~Schmidt \& H.~Lipson}]{schmidt-lipson2009}
Schmidt, Michael \& Hod Lipson. 2009. Distilling free-form natural laws from
  experimental data. {\em Science\/} 324, 81--85.

\bibitem[{Sch\"{o}ner(1990)G.~Sch\"{o}ner}]{schoener1990}
Sch\"{o}ner, Gregor. 1990. A dynamic theory of coordination of discrete
  movements. {\em Biological Cybernetics\/} 63(4), 257--210.

\bibitem[{Sch\"{o}ner(2020)G.~Sch\"{o}ner}]{schoener2020}
Sch\"{o}ner, Gregor. 2020. The dynamics of neural populations capture the laws
  of the mind. {\em Topics in Cognitive Science\/} 12, 1257--1271.

\bibitem[{Sch\"{o}ner, Spencer \& {The DFT Research Group}(2016)G.~Sch\"{o}ner,
  J.~P. Spencer \& {The DFT Research Group}}]{schoener-etal2016}
Sch\"{o}ner, Gregor, John~P. Spencer \& {The DFT Research Group}. 2016. {\em
  Dynamic Thinking: A Primer on Dynamic Field Theory\/}. Oxford: Oxford
  University Press.

\bibitem[{Schr\"{o}dinger(1944)E.~Schr\"{o}dinger}]{schrodinger1944}
Schr\"{o}dinger, Erwin. 1944. {\em What is Life? The Physical Aspect of the
  Living Cell\/}. Cambridge: Cambridge University Press.

\bibitem[{Senkowski \& Engel(2024)D.~Senkowski \& A.~K.
  Engel}]{senkowski-engel2024}
Senkowski, Daniel \& Andreas~K. Engel. 2024. Multi-timescale neural dynamics
  for multisensory integration. {\em Nature Reviews Neuroscience\/} 1--18.

\bibitem[{{\v{S}imko} \& Cummins(2010)J.~{\v{S}imko} \&
  F.~Cummins}]{simko-cummins2010}
{\v{S}imko}, Juraj \& Fred Cummins. 2010. Embodied task dynamics. {\em
  Psychological Review\/} 117(4), 1229--12246.

\bibitem[{Smolensky(1988)P.~Smolensky}]{smolensky1988}
Smolensky, Paul. 1988. On the proper treatment of connectionism. {\em
  Behavioral and Brain Sciences\/} 11(1), 1--74.

\bibitem[{Sorensen \& Gafos(2016)T.~Sorensen \& A.~I.
  Gafos}]{sorensen-gafos2016}
Sorensen, Tanner \& Adamantios~I. Gafos. 2016. The gesture as an autonomous
  nonlinear dynamical system. {\em Ecological Psychology\/} 28(4), 188--215.

\bibitem[{Sorensen \& Gafos(2023)T.~Sorensen \& A.~I.
  Gafos}]{sorensen-gafos2023}
Sorensen, Tanner \& Adamantios~I. Gafos. 2023. The relation between gestures
  and kinematics. In Florian Breit, Bert Botma, Marijn {van 't Veer} \& Marc
  {van Oostendorp} (eds.) {\em Primitives of Phonological Structure\/},
  251--279, Oxford: Oxford University Press.

\bibitem[{Spivey(2007)M.~Spivey}]{spivey2007}
Spivey, Michael. 2007. {\em The Continuity of Mind\/}. Oxford: Oxford
  University Press.

\bibitem[{Stern \& Shaw(2023)M.~C. Stern \& J.~A. Shaw}]{stern-shaw2023}
Stern, Michael~C. \& Jason~A. Shaw. 2023. Neural inhibition during speech
  planning contributes to contrastive hyperarticulation. {\em Journal of Memory
  and Language\/} 132(104443), 1--16.

\bibitem[{Stern \& Shaw(2024)M.~C. Stern \& J.~A. Shaw}]{stern-shaw2024}
Stern, Michael~C. \& Jason~A. Shaw. 2024. Towards a minimal dynamics for
  gestures: A law relating velocity and position. {\em Proc. {ISSP} 2024 --
  13th International Seminar on Speech Production\/} 262--265.

\bibitem[{Strogatz(2015)S.~H. Strogatz}]{strogatz2015}
Strogatz, Steven~H. 2015. {\em Nonlinear Dynamics and Chaos: With Applications
  to Physics, Biology, Chemistry, and Engineering\/}. Philadelphia, PA:
  Westview Press, second edn.

\bibitem[{Suprunenko, Clemson \& Stefanovska(2013)Y.~F. Suprunenko, P.~T.
  Clemson \& A.~Stefanovska}]{supruneko-etal2013}
Suprunenko, Yevhen~F., Philip~T. Clemson \& Aneta Stefanovska. 2013.
  Chronotaxic systems: A new class of self-sustained nonautonomous oscillators.
  {\em Physical Review Letters\/} 111(2), 024101.

\bibitem[{Tilsen(2009)S.~Tilsen}]{tilsen2009}
Tilsen, Sam. 2009. Multitimescale dynamical interactions between speech rhythm
  and gesture. {\em Cognitive Science\/} 33(5), 839--879.

\bibitem[{Tilsen(2016)S.~Tilsen}]{tilsen2016}
Tilsen, Sam. 2016. Selection and coordination: The articulatory basis for the
  emergence of phonological structure. {\em Journal of Phonetics\/} 55, 53--77.

\bibitem[{Tilsen(2018)S.~Tilsen}]{tilsen2018}
Tilsen, Sam. 2018. Three mechanisms for modeling articulation: selection,
  coordination, and intention. {\em Cornell Working Papers in Phonetics and
  Phonology\/} 1--49.

\bibitem[{Tilsen(2019{\natexlab{a}})S.~Tilsen}]{tilsen2019}
Tilsen, Sam. 2019{\natexlab{a}}. Motoric mechanisms for the emergence of
  non-local phonological patterns. {\em Frontiers in Psychology\/} 10(2143),
  1--25.

\bibitem[{Tilsen(2019{\natexlab{b}})S.~Tilsen}]{tilsen2019b}
Tilsen, Sam. 2019{\natexlab{b}}. {\em Syntax with oscillators and energy
  levels\/}. Berlin: Language Science Press.

\bibitem[{Tilsen(2020)S.~Tilsen}]{tilsen2020}
Tilsen, Sam. 2020. A different view of gestural activation: Learning gestural
  parameters and activations with an {RNN}. {\em Cornell Working Papers in
  Phonetics and Phonology\/} 1--49.

\bibitem[{Tilsen(2022)S.~Tilsen}]{tilsen2022}
Tilsen, Sam. 2022. An informal logic of feedback-based temporal control. {\em
  Frontiers in Human Neuroscience\/} 16(851991), 1--29.

\bibitem[{Turing(1950)A.~M. Turing}]{turing1950}
Turing, Alan~M. 1950. Computing machinery and intelligence. {\em Mind\/}
  59(236), 433--460.

\bibitem[{Turk \& Shattuck-Hufnagel(2020)A.~Turk \&
  S.~Shattuck-Hufnagel}]{turk-shattuck-hufnagel2020}
Turk, Alice \& Stefanie Shattuck-Hufnagel. 2020. {\em Speech Timing:
  Implications for Theories of Phonology, Phonetics, and Speech Motor
  Control\/}. Oxford: Oxford University Press.

\bibitem[{{van Gelder}(1998)T.~{van Gelder}}]{vangelder1998}
{van Gelder}, Timothy. 1998. The dynamical hypothesis in cognitive science.
  {\em Behavioral and Brain Sciences\/} 21(1), 615--665.

\bibitem[{Virtanen, Gommers, Oliphant et~al.(2020)P.~Virtanen, R.~Gommers,
  T.~E. Oliphant, M.~Haberland, T.~Reddy, D.~Cournapeau, E.~Burovski,
  P.~Peterson, W.~Weckesser, J.~Bright, S.~J. {van der Walt}, M.~Brett,
  J.~Wilson, K.~J. Millman, N.~Mayorov, A.~R.~J. Nelson, E.~Jones, R.~Kern,
  E.~Larson, C.~J. Carey, {\.I}.~Polat, Y.~Feng, E.~W. Moore, J.~{VanderPlas},
  D.~Laxalde, J.~Perktold, R.~Cimrman, I.~Henriksen, E.~A. Quintero, C.~R.
  Harris, A.~M. Archibald, A.~H. Ribeiro, F.~Pedregosa, P.~{van Mulbregt} \&
  {SciPy 1.0 Contributors}}]{SciPy-NMeth2020}
Virtanen, Pauli, Ralf Gommers, Travis~E. Oliphant, Matt Haberland, Tyler Reddy,
  David Cournapeau, Evgeni Burovski, Pearu Peterson, Warren Weckesser, Jonathan
  Bright, St{\'e}fan~J. {van der Walt}, Matthew Brett, Joshua Wilson, K.~Jarrod
  Millman, Nikolay Mayorov, Andrew R.~J. Nelson, Eric Jones, Robert Kern, Eric
  Larson, C~J Carey, {\.I}lhan Polat, Yu~Feng, Eric~W. Moore, Jake
  {VanderPlas}, Denis Laxalde, Josef Perktold, Robert Cimrman, Ian Henriksen,
  E.~A. Quintero, Charles~R. Harris, Anne~M. Archibald, Ant{\^o}nio~H. Ribeiro,
  Fabian Pedregosa, Paul {van Mulbregt} \& {SciPy 1.0 Contributors}. 2020.
  {{SciPy} 1.0: Fundamental Algorithms for Scientific Computing in Python}.
  {\em Nature Methods\/} 17, 261--272.

\bibitem[{Westbury(1994)J.~R. Westbury}]{westbury1994}
Westbury, John~R. 1994. {\em X-Ray Microbeam Speech Production Database User's
  Handbook\/}. Madison, WI: Waisman Center.

\end{thebibliography}
\end{document}